# ROSE: A Neurocomputational Architecture for Syntax

Elliot Murphy[1,2]

**Affiliations**: 1. Vivian L. Smith Department of Neurosurgery, McGovern Medical School, UTHealth, Houston, TX, USA

2. Texas Institute for Restorative Neurotechnologies, UTHealth, Houston, TX, USA

**Correspondence**: elliot.murphy@uth.tmc.edu

**Abstract**: A comprehensive model of natural language processing in the brain must accommodate four components: representations, operations, structures and encoding. It further requires a principled account of how these different components mechanistically, and causally, relate to each another. While previous models have isolated regions of interest for structure-building and lexical access, and have utilized specific neural recording measures to expose possible signatures of syntax, many gaps remain with respect to bridging distinct scales of analysis that map onto these four components. By expanding existing accounts of how neural oscillations can index various linguistic processes, this article proposes a neurocomputational architecture for syntax, termed the ROSE model (Representation, Operation, Structure, Encoding). Under ROSE, the basic data structures of syntax are atomic features, types of mental representations (R), and are coded at the single-unit and ensemble level. Elementary computations (O) that transform these units into manipulable objects accessible to subsequent structure-building levels are coded via high frequency broadband γ activity. Low frequency synchronization and cross-frequency coupling code for recursive categorial inferences (S). Distinct forms of low frequency coupling and phase-amplitude coupling (δ-θ coupling via pSTS-IFG; θ-γ coupling via IFG to conceptual hubs in lateral and ventral temporal cortex) then encode these structures onto distinct workspaces (E). Causally connecting R to O is spike-phase/LFP coupling; connecting O to S is phase-amplitude coupling; connecting S to E is a system of frontotemporal traveling oscillations; connecting E back to lower levels is low-frequency phase resetting of spike-LFP coupling. This compositional neural code has important implications for algorithmic accounts, since it makes concrete predictions for the appropriate level of study for psycholinguistic parsing models. ROSE is reliant on neurophysiologically plausible mechanisms, is supported at all four levels by a range of recent empirical research, and provides an anatomically precise and falsifiable grounding for the basic property of natural language syntax: hierarchical, recursive structure-building.

**Keywords**: Syntax; neural oscillations; ROSE; delta; traveling waves

Modern linguistics has arrived at the general assumption that the human mind/brain applies a set of syntactic rules to recursively combine linguistic units into larger objects, deriving an unbounded array of hierarchically structured expressions, with humans inferring sentence meaning based on syntactic configuration (Chomsky, 1957; Everaert et al., 2015; Hagoort, 2023; Mukherji, 2022). Despite some recent claims that "there are no traits present in humans and absent in other animals that in isolation explain our species' superior cognitive performance", and that humans are simply "flexible cognitive allrounders" (Laland and Seed, 2021, p. 689), amongst other claims that language cannot even be given a biological account (Smit, 2022), many linguists maintain, as I will here, that this capacity for constructing hierarchical syntactic objects and assigning them a categorized, labeled identity is human-specific, even if the generic facility for recursion might be shared with other species (Liao et al., 2022). Furthermore, this capacity for hierarchical recursion has been linked to human-specific cognitive superiority (Dehaene et al., 2022; Hauser and Watumull, 2017). Certain aspects of language can only be captured via the postulation of structures with relations between their elements (Everaert et al., 2015), motivating a qualitative computational distinction between human and non-human psychology, in contrast to frameworks invoking "allrounders".

Meanwhile, an emerging consensus in neuroscience is that complex behavior and cognition rely on coordinated interactions between brain regions (Miller and Wilson, 2008), with phase synchronization being a major candidate for a mechanism implementing this coordination, by gating information transmission (Bressler and Kelso, 2001). Yet, unlike for models of attention and working memory, there is a current absence of oscillatory phase coding in models of natural language.

The central question that will occupy this article concerns whether we can begin to draw causal, mechanistic connections between theoretical linguistics and the brain sciences. I will begin by reviewing some current accounts that map syntactic operations onto neural systems, before proposing a novel neurocomputational architecture for syntax. After progressing to the theme of causal evidence for this architecture, I will evaluate some open questions and future directions, before concluding with a focus on testability.

# Cracking the Neural Code for Syntax

The past decade has seen the emergence of low frequency phase coherence as a feasible index of hierarchical syntactic structure building. This measure refers to whether two or more brain regions have similar oscillatory activity based on phase consistency. This has been associated chiefly with rule-based chunking rather than morphophonological or semantic features of phrases, typically utilizing extra-cranial recording (Lu et al., 2022a, b). Concurrently, recent work has examined local cortical processing with intracranial recordings, often focusing instead on high frequency power changes. This work has exposed signatures for basic linguistic operations and conceptual categories (Murphy et al., 2022c; Woolnough et al., 2022a, c). How these two distinct recording scales can be combined into a coherent model of natural language syntax and cortical computation has yet to be addressed.

One way to explore this issue is through the lens of an existing neurocomputational model of syntax (Murphy, 2020a). The combinatorial power of language is here indexed via various oscillatory interactions such as forms of cross-frequency coupling. In Murphy (2020a), empirical and conceptual motivations are presented to defend the idea that δ-θ inter-regional phase-amplitude coupling constructs multiple sets of syntactic and semantic features. This occurs when the phase of δ is synchronized with the amplitude of θ. δ represents supraordinate syntactic categories, and θ represents feature-bundles generated via lexical access. Phase-resetting (Harris, 2023) of this mechanism alongside concurrent encoding/storage of its products in workspaces before a newly-generated δ-θ complex is created permits a facility for recursive self-call. Distinct β (inferior frontal cortex) and γ sources (cross-cortical sites of conceptual storage; see below) are also coupled, respectively, with δ and θ (e.g., θ-γ phase-amplitude coupling) for syntactic prediction and conceptual binding.

Traveling oscillations will also be relevant for our discussion (Muller, 2018; Murphy, 2018; Woolnough et al., 2022b). These are oscillations which migrate in phase coherence across the brain, whereby the spiking of neural clusters is coordinated not just across two fixed points but across a particular extended path. Traveling oscillations support brain connectivity and function (Zhang et al., 2018). In Murphy (2020a), δ waves cycle across left posterior temporal and inferior frontal cortices, building up the syntactic workspace as successive θ-complexes are synchronized with δ. Traveling δ waves are assumed to be responsible for patterning spiking from single- to multi-unit lexical structures in each δ cycle. As such, δ would coordinate phrasal construction while θ-γ interactions would support the representational construction of linguistic feature-bundles. δ-γ coupling is involved in fluid intelligence (Gągol et al., 2018) and linguistic phrase composition via low γ activity detected with scalp EEG (Brennan and Martin, 2020), whereby δ embeds cross-cortical γ rhythms depending on the cortical areas needed for the particular task, i.e., geometric reasoning, visual processing. Murphy (2018, 2020a) proposes that δ-γ coupling may be a generic combinatorial process, combining representations from within and across domains. Brennan and Martin (2020) discovered an increasing scale of δ-θ and δ-γ coupling beginning at the point of a word completing a single phrase, progressing into multi-phrase units, supporting the foundations of the current model. The finding of increasing θ and β activity in MEG during incremental sentence processing also supports these assumptions (Bastiaansen et al., 2010).

Lastly, the basic aspects of the oscillatory dynamics of language result from genetic guidance, and a confident list of candidate genes for this guidance can be posited (Murphy and Benítez-Burraco, 2018); these candidate genes map onto specific aspects of brain function, particularly onto neurotransmitter function, and through dopaminergic, GABAergic and glutamatergic synapses.

Below, I will expand on this model in ways that help address some important recent critiques of it (e.g., Kazanina and Tavano, 2022; Leivada, 2017; Martorell, 2021; Mondal, 2022; Mukherji, 2022) pertaining to the separability of syntax from semantics, issues of precise spatiotemporal details, the possiblility that oscillations are simply 'emergent', the apparent non-uniqueness to humans of oscillatory binding mechanisms, and causal evidence. Reasons to focus on this model can be found via some recent empirical support for it: Dekydtspotter et al. (2023) test the model in Murphy (2020a) and find support for syntactic processing load effects in β from EEG recordings of anaphora resolution in successive *wh*-movement in native and nonnative speakers of French.

Aiding with initial direction, Ding (2022) poses perhaps the most important question for the neurobiology of syntax: What type of neural activity, such as low-frequency activity, high γ power, or spiking rate, can track very long or very brief phrases? I will try to answer this question here in the following way, associating roughly the respective scales of recording and neurobiological organization with particular levels of linguistic complexity: low frequency activity (*structures*), high γ power (*operations*), spiking rate (*basic units/representations*). I will argue that present models of syntax do not adjudicate between appropriate levels of linguistic complexity and scales of neural complexity (unlike, for example, neural models of speech processing).

## Current Neural Models of Syntax

Building on past insights, new brain models of syntax should seek not only to incrementally add to existing knowledge and assumptions – they should also occasionally be "disruptive" (Kozlov, 2023; Park et al., 2023) with respect to existing pre-conceptions and disciplinary boundaries. While traditional models of working memory (e.g., Baddeley and Hitch, 1974) have been replaced by neurophysiologically sophisticated models (e.g., Lisman and Jensen, 2013), within linguistics the traditional localizationist models have yet to be supplanted in a similar way.

A comprehensive model of natural language processing in the brain must accommodate four components: *representations*, *operations*, *structures*, and their *encoding* and storage in short-term memory. Previous accounts of syntax in the brain have focused mostly, or exclusively, on only a subset of these components. For example, Friederici (2017) entertains a model of linguistic structures (S) and encoding (E), but omits details about the electrophysiological or neurophysiological basis of representations (R) and operations (O).

When summarizing previous models below, I stress that I will be evaluating these purely with respect to how they address the implementational basis of R, O, S and E. Each of these following models have great strengths with respect to how they frame various components of syntax (be that at the elementary structure-building level, or the complex sentential level), but my goal here is not to evaluate these models against each other; rather, it is to address how they deal with the issue of resolving appropriate neural mechanisms for the basic computational components of syntax.

Beginning with the model in Murphy (2020a), outlined above: This took into account O and S, there was no focus on R or how distinct components of linguistic structure are related and encoded (E).

Matchin and Hickok (2020) take into account the localization of O and S, with less emphasis on E. Although these authors present extensive evidence to distinguish the computational roles of posterior temporal cortex (lexical-syntactic structures) and inferior frontal cortex (morphosyntactic linearization), once we have successfully localized these computational distinctions, this is effectively the end of the road for the localizationist project.

Krauska and Lau (2022) take full consideration of the appropriate computational-level resolution for R, O, S and E, and provide a clear algorithmic-level model of syntactic processing, but the implementation-level resolution for each of these components remains unresolved.

Following the centralization of labeling in the psycholinguistics of syntax proposed in recent work (Adger, 2017; Chomsky et al., 2019; Murphy, 2015a, b), Goucha et al. (2017) focus on the importance of labeling in neurolinguistic models of syntax. They keep to network-level tractographic concerns of motivating how frontotemporal regions interface to generate S and E, with no relations to R or O. Although Goucha et al. (2017) defend a clear picture of hierarchy-category relations across frontotemporal sites, the implementation of this keeps to discussion of "cross-talk" between regions. In addition, the central theme of the model is that "labeling would be anatomically […] enabled by the arcuate fascicle", but what precisely this enabling would look like is not explored. Nevertheless, it remains well-motivated to focus on structure-building via labeling, since this operation feeds recursivity and provides a uniform account of various syntactic phenomena, such as traditional islands and locality-of-movement effects (Bošković, 2021).

The model in Pylkkänen (2019) provides a computationally explicit depiction of the general spatiotemporal dynamics of structure/meaning inferences, taking O and S as the focal points of discussion, but to the exclusion of R and E.

Other models propose hypotheses about O and S, but the causal and mechanistic connection between these levels is not addressed. Some models that take neurophysiological details seriously nevertheless fall short with respect to fleshing out computational details for neural activity scales; e.g., in Martin (2020) γ activity is associated with "the retrieval of memory-based linguistic representations by minimal or thresholded acoustic cues" (but see Kaushik and Martin, 2022, for new directions here). What is needed is not just an account of R, O, S and E, but also a principled means to *link* these components.

Recent research has shown that the human brain and autoregressive deep language models both engage in *continuous next-word prediction* before word onset (Goldstein et al., 2022), but this literature has focused almost exclusively on single-element, lexical statistics and prediction, which many linguists would argue do not form the core basis of human language (see also Oota et al., 2022).

Empirical issues also remain. The claim that the basic structure-building operation MERGE "appears to be localized in a very confined region" (Friederici and Singer, 2015, pp. 334-335), BA 44, does not align with intracranial recordings (Murphy et al., 2022e) and lesion data (Matchin et al., 2020, 2022). Chronic Broca's aphasia is associated with damage to both Broca's and Wernicke's areas (Fridriksson et al., 2015). Young et al. (2021) review a range of existing results from intracranial cortical stimulation mapping and showed how left inferior frontal regions do indeed play a role in syntax-semantics, but likely in complex sentence processing, rather than in elementary structure-building.

Turning to other models, Hagoort's (2005, 2013) Memory-Unification-Control (MUC) model takes into consideration O and E, but R and S appear only via localizationist assumptions of mappings to gross anatomical regions, rather than being assembled bottom-up via specific neurophysiological mechanisms. MUC assumes that the Memory component constitutes "the only language-specific component of the model" (Hagoort, 2013), with Unification being directed by lexico-semantic properties and syntactic templates inherent to Memory. Unification is also assumed to take place in frontal cortex, however inferior frontal cortex seems to be modulated more specifically by unification demands, or the *use* of structure for a given task (Murphy et al., 2022e). Hagoort has qualified this stance by claiming that Broca's area forms only the major node in the Unification network, and that it is the *interactions* between Broca's area and posterior temporal gyrus that are crucial (see also Baggio, 2018; Baggio and Hagoort, 2011), opening up further questions about the relative involvement (Murphy et al., 2022e), and causal involvement (Matchin et al., 2022), of frontal cortex in structure-building. Other claims about inferior frontal cortex ("Activity in LIFC is presumably relatively insensitive to the onset and offset times of the stimuli, and is rather a self-sustaining state which is relatively unaffected by trial-to-trial variation"; Hagoort, 2013) are incompatible with intracranial recordings of rapid frontal responses to linguistic stimuli of varying types, including the integration of lexical, phrasal and sentential meaning (Woolnough et al., 2021, 2022a, 2023).

With respect to models of semantics more broadly, the Controlled Semantic Cognition account (Lambon Ralph et al., 2017) is split into a representation component and a control component, for storing and manipulating semantic information during verbal and non-verbal tasks. While this provides a clear algorithmic picture for accessing and manipulating elementary semantic representations, and provides a clear localizationist topography, the model currently lacks any relation to specific levels of neurophysiological complexity. This critique also applies to even the most recent reviews of the neurocomputational basis of semantic memory (Frisby et al., 2023), despite algorithmic models of semantic memory becoming more sophisticated (e.g., distinct types of contiguous or dispersed codes).

The work of Dehaene and colleagues (Dehaene et al., 2015, 2022) has perhaps come the closest to relating scales of neural activity to language. Yet, even here focus is placed mostly on R and O with no linking hypotheses between them ("The precise neural mechanisms of chunk formation remain unknown, and are probably widely distributed in multiple cortical areas"; Dehaene et al., 2015, p. 6). Under Dehaene's framework, there are no principled delimitations as to where, and why, we would detect signatures of syntax, that might be built from independent assumptions concerning how the brain might encode basic units of information, or parse complex tree structures in real time. The frameworks in Dehaene et al. (2015, 2022) seem equally open to discovering cellular 'barcodes' for syntactic tree-structure composition as they are open to discovering more global signatures, which is certainly sound experimental practice but it also renders it more difficult to motivate any specific neural architecture tying different levels of linguistic complexity together (but see Desbordes et al., 2023 for new directions mapping out the neural geometry of linguisitc composition). Although this is admittedly more of a critique of the existing state of the field, I will claim below that a more systematic means of constraining neural models of syntax exists.

Lastly, Poeppel and Idsardi (2022) maintain that neurolinguistic models promulgating associationist frameworks to brain-language mappings will fall short of genuine explanation. They further claim that words should be seen as "a conjunction of form and concept" (where forms can be speech, sound, sign, Braille). This neatly accounts for a range of orthographic and auditory forms, but we should also qualify that form-concept conjunctions are not the full story either, with silent syntactic and inert elements standing on less clear ground. In addition, Poeppel and Idsardi's (2022) form-concept conjunction account does not pertain to the many ways that lexicalization of a concept seems to alter it, to imbue it with elements of aiteational formatting, or teleological or functional identity, in ways that deny many common nouns simple definitions (Chomsky, 2000; Gotham, 2016). The ultimate cognitive or neurobiological reason for why lexicalization seems to place concepts on a different 'plane' is unclear (see Hagoort, 2023 and Pietroski 2018 for related discussion).

## ROSE

As noted, any comprehensive account of syntax needs to accommodate *Representations* (atomic units of computation), *Operations* (manipulation and transformation of representations into a novel format), *Structures* (arrangements of representations into hierarchical relations with independent identities not reliant purely on the properties of individual units) and *Encoding* (the real-time maintenance of structures, and the 'shielding' of their identity against decay and transformation). As such, I will defend what I will term the ROSE model.

Below is an outline of the basic scales of analysis that map onto each component of the model, which I will then motivate throughout the remainder of this article (see Figure 1 for a schema of components with their associated causal/mechanistic links).

**Representation-Operation-Structure-Encoding (ROSE)**

**R**: Single-unit encoding of conceptual features and formal syntactic features. This level involves a cellular barcode for distinct features that compose into syntactic objects coherently bound by high γ at O. It also involves vector codes for ensembles hosting features common to objects represented at O and that are ultimately coordinated by S.

**O**: High γ sensorimotor transformation into a lexicalized object (core network nodes: mid-fusiform gyrus, orbitofrontal cortex, middle temporal gyrus) accessible to δ/θ phase-locking. This level can implement the semantic composition of language-specific concepts (minimal phrase schemes) that coordinate the firing of R units. High γ activates assemblies of distinct units hosting the barcode or vector code for units $R_1 \ldots R_n$ that compose into feature-bundles.

**S**: A low frequency neural program for generating structural inferences over O. δ-θ phase-amplitude coupling (posterior superior temporal sulcus to inferior frontal gyrus) for categorial inferences modulating the representation of feature-bundles in θ-γ (inferior frontal gyrus to cross-modular hubs).

**E**: Local and global workspaces for bottom-up lexical memory and top-down hierarchical memory. Traveling waves implement δ-θ coupling for hierarchical memory, and θ-γ coupling for lexical memory. α power codes for workspace 'disruption' (posterior temporal, inferior parietal cortex). β power coding for syntactic predictions (inferior frontal cortex).

Under ROSE, high γ activity reflects operations like lexical selection from a set in Broca's area, or orthographic-to-phonological conversions in inferior parietal cortices – core operations of language and its interfaces, but these do not appear to directly index units of computation (R).

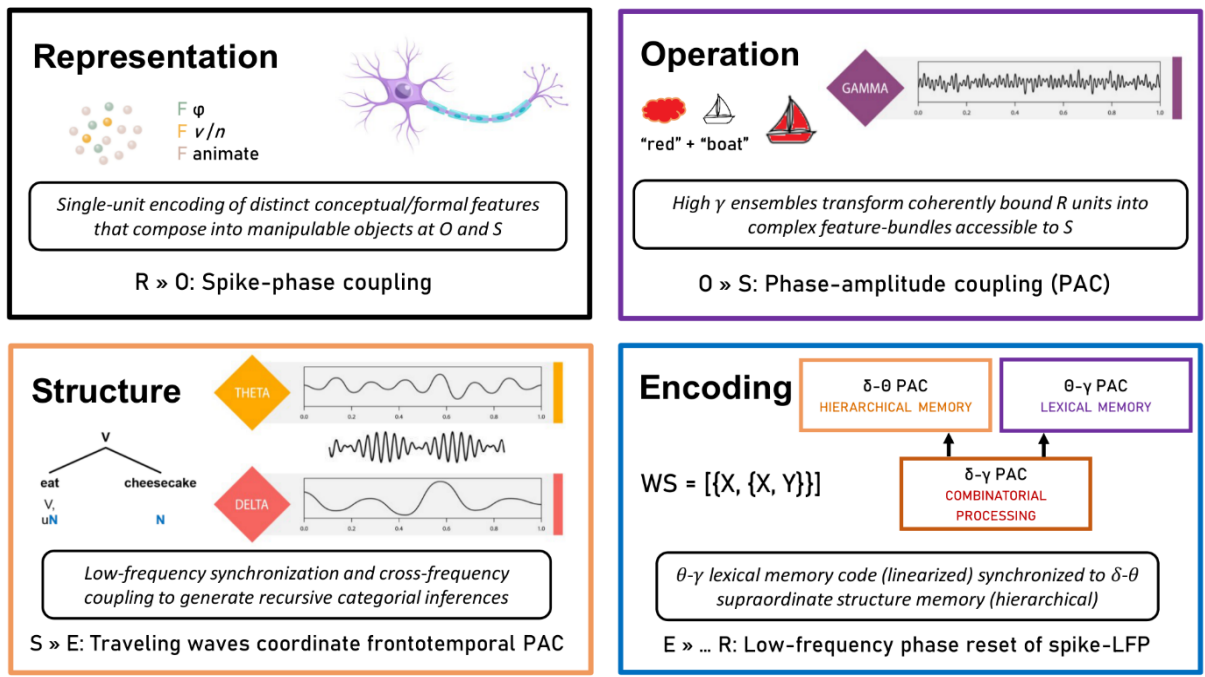

**Figure 1:** Basic components of the ROSE model. Each box contains a graphical depiction of the relevant organizational scale (top), a description of how each scale relates to the relevant component of language (middle, boxed text), and the putative mechanistic link between components (bottom text). Components are named representation (R), operation (O), structure (S), encoding (E). The colored boxes in the Encoding component (hierarchical memory, lexical memory) correspond to the colors of the level of linguistic representation they are associated with (i.e., S for hierarchical memory, O for lexical memory).

Some researchers will claim that the brain does not recognize constructs like 'tense' or 'pronominal binding' because they cannot detect a signature of these constructs at a particular resolution. However, many such anxieties and issues with current models of language in the brain can be overcome if we more carefully entertain appropriate scales of neural complexity and signal processing. Instead of viewing each component of the model separately, successive levels of syntactic composition under ROSE follow mechanistically from the next, constituting a full-scale model with principled and falsifiable (Kleiner and Hoel, 2021) connections. The most advanced neurocomputational models of language processing (reviewed in Hale et al., 2022) rely on decoding BOLD responses and scalp EEG signatures, dissociating hierarchical from lexical-related signatures across specific time windows, but there is no clear demarcation between the components in ROSE with respect to their levels of neural organization. Many such models are also extensional in character, as opposed to being concerned with the operations and representations that are needed to arrive at the output of language processing (but see Brennan et al., 2020, who centralize psycholinguistic processes and algorithmic steps). Other models make lexicalist distinctions between lexical node representations and higher-order structure representations in posterior temporal cortex (Wilson et al., 2018), but clear-cut localizationist distinctions between lexicality/structure seem unmotivated (Murphy et al., 2022e). Taking again Hagoort's MUC model to illustrate this argument, there is no clear neurophysiological transition across M, U and C, and all three components lack a specific motivation in terms of a scale of neurobiological complexity. For example, it is unclear which scale of neural complexity is optimal across M, U and C.

I will now turn to motivating each component of ROSE, before presenting converging causal evidence, and concluding with a discussion of future prospects. Before doing so, I note here that the discussion below should be seen as being aligned with assumptions about language processing that try and integrate computation-level accounts with algorithm-level accounts, such as the 'one-system' contention (Lewis and Phillips, 2015) that grammatical theories and language processing models describe the same cognitive system. This is evidenced by grammar-parser misalignments occurring as a consequence of limitations in domain-general systems such as memory access and control mechanisms, and is also evidenced by the convergence between online and offline responses to grammatical anomalies (Sprouse and Almeida, 2013). While I do not aim here to adjudicate between rival parsing models (e.g., various flavours of minimalist parsers), I note that one cannot even begin to robustly test distinct parsing models in a neurophysiologically plausible way if the appropriate level of neural organization is not first arrived at for the specific process under consideration. The architecture of ROSE is precisely that: an architecture, compatible with a range of psycholinguistic accounts.

**Representation**

Recent work at the single-unit level in humans has elucidated place and time cells, face cells, stubby shape cells, spiky shape cells, and cells sensitive to animacy (Bao et al., 2020). What will single-unit resolution recordings for language look like? Under ROSE, small ensembles of temporarily cooperating neurons, some of which are highly specialized for particular lexico-semantic feature types (e.g., φ, Q), collectively represent atomic features that operate as feature-bundles. Semantic features of lexical items likely "combine sparse codes with the flexibility and combinatorial richness of assembly codes" (Friederici and Singer, 2015), with semantically related items exhibiting overlapping single-unit responses across medial temporal and lateral posterior middle temporal cortices, depending on approximation to canonical categories that these regions seem sensitive to. For instance, posterior middle temporal cortex may serve language-specific representational features, such as abstract word meaning, and may be involved in lexicalization of non-linguistic concepts (intracranial evidence for the unique involvement of posterior temporal cortex in abstract word representations – often framed and domain-specific – supports this view; Murphy et al., 2022d). Verguts (2017) plausibly speculates that different neural populations which have their spiking patterns modulated by γ rhythms are, "presumably, cognitive representations". The full spread of concept-specific activation profiles for linguistic features remains to be comprehensively mapped, in particular those pertaining to features with grammatical/formal status, but new research is beginning to emerge here; for example, the temporal features of lexical items appear to be partially subserved by time perception circuits in parietal cortex (Johari et al., 2023).

Syntactic features are assumed here to be represented much like other long-term memory representations. Recent work suggests that single cells in human medial temporal lobe can shift their concept-related selectivity based on task demands while other cells are less susceptible to being re-coded (Donoghue et al., 2023). For syntactic features, it is possible that a similar kind of mismatch of fixed vs. flexible cells exists, and that this varies across feature type (e.g., formal, semantic, categorial). To elaborate on this point further, and to stress caution when assigning concept-related or feature-related neuronal profiles, consider how grid cells had their representational profile expand from spatial navigation to auditory navigation to conceptual navigation (in brief, navigation in all its guises) (Mok and Love, 2019). In contrast, face cells have enjoyed the opposite journey, shifting from concept cell status to recently being implicated in navigation (Khandhadia et al., 2023). Similar journeys may soon await syntax-related and theta-role-related cells.

The major, relevant activity scales for the R and O components of ROSE are as follows:

> **SUA (single-unit activity)**: Extracellularly-recorded activity of an individual neuron.
>
> **MUA (multi-unit activity)**: Multiple neurons, between ~50–300 microns (Buzsáki, 2004).
>
> **ESA (envelope of spiking activity)**: The envelope of MUA (Ahmadi et al., 2021).
>
> **LFP (local field potential)**: Mesoscopic, large number of neurons, from ~250μ to several millimeters (Buzsáki et al., 2012).

Before moving to the specifics of the current model, I stress that the notion of representation entertained here is synonymous with atomic feature or unit: I am not concerned with, for example, notions of representation found with formulations such as 'the syntactic representation of multiple *wh*-movemenet', or 'the representation of long-distance pronominal binding relations'. I am concerned with elementary representations – there are also *representations of structure* or *workspace representations*, but to avoid terminological confusion I will simply refer to these below as S and E.

*Basic Units*

The atoms of linguistic computation are commonly assumed to be bundles of features or functional categories (Chomsky, 1967, 1974; Ramchand and Svenonius, 2014). However, linguists within, for example, the Distributed Morphology tradition have abandoned hope in a coherent notion of 'word'; there are simply groups of features coding for conceptual roots, derivations and inflections (Gwilliams, 2020). Morphology and syntax are part of the same structure-building system, hence why we can generate words that, structurally, look like 'mini sentences', like *anti-institutionalization*. In polysynthetic languages, a single word can be composed of many productive morphemes, representing complex meanings. Although further specifics will not concern us, ROSE can accommodate (i) sets of syntactic atoms (R), (ii) mapping rules between syntax and concepts (O), and (iii) mapping rules between syntax and form (i.e., production) (Preminger, 2021). ROSE therefore permits the single-unit or assembly-based representations of distinct features from these domains, as opposed to single-unit encoding of a 'lexical item' and 'word'. Stored linguistic knowledge can involve the mapping from a semantic feature, to a syntactic feature, and to a phonological feature, and this process is hypothesized to take place at the O level (see below).

Emerging evidence for this framework comes from Dijksterhuis et al. (2022), in which pronouns referring to previous discourse entities reactivate single cells in the human medial temporal lobe. Other work points to single-neuron representations of ambiguous words (specifically, homonym), also in the medial temporal lobe (Samimizad et al., 2022). Relatedly, Cai et al. (2022) conducted single-neuron recordings in five patients undergoing planned intraoperative interventions, and exposed them to auditory sentence recordings. They found that many neurons in lateral inferior frontal cortex and lateral middle temporal cortex responded significantly to a variety of lexical features during comprehension (lexico-semantic features and thematic roles), and some were also engaged in accurate next-word prediction.

These recent studies all point to the existence of coordinated assemblies of neurons that encode a variety of lexico-semantic features. This is likely done in a mostly redundant format, although the range of inter-featural combinations available to a given representation (e.g., syntactic feature X can combine with sets Y and Z, but feature P can only combine with set Y) may relate to whether an atomic feature is sparsely (cell-specific) or redundantly encoded.

Other work from Nelson et al. (2022) discovered preferential activity for content words over function words in the anterior temporal lobe, independent of word length and word position in the sentence. Indeed, anterior temporal regions (along with ventromedial prefrontal cortex) appear to be involved in conceptual combination in basic phrases, but not syntactic combination (Pylkkänen, 2019), important for semantic hub-based operations. More generally, concept cells that preferentially activate for stimuli that occupy a certain position in some categorial axis (e.g., animate/inanimate, spiky/stubby; Bao et al., 2020), seem to support declarative memory formation. Together with the more established finding that high frequency ripple oscillations (~150-250 Hz) and fast ripple (250-500 Hz) activity is modulated by human memory encoding and recall across the cortex (Kucewicz, 2014), these new results point to the involvement of local cortical circuits and individual neurons encoding basic atomic features.

Lastly, an extensive study of lexical access in epilepsy patients (via common object naming) has revealed another strong candidate for a lexico-semantic feature hub, the mid-fusiform gyrus and surrounding ventral temporal cortex, which are strongly associated with post-operative naming decline (Snyder et al., 2023). Importantly, Snyder and colleagues note that these loci coincide with the sites of susceptibility artifacts during echo-planar imaging, potentially explaining why this region has been previously underappreciated with respect to its role in lexical access (both in the clinical and neurolinguistic worlds).

All together, these recent results indicate the existence of previously under-appreciated scales of neural complexity and cortical sites that code for basic R units.

*Mapping from R to O*

How can we relate these ideas to O, the component of ROSE responsible for manipulating and combining R units rather than simply accessing them? Local field potentials (LFPs) reflect the common synaptic activity of a population of neighboring neurons (or the weighted sum of the aggregate activity of a group of neurons, reflecting mostly postsynaptic activity) (Buzsáki et al., 2012; Freeman, 1975; Jansen et al., 2015; Łęski et al., 2013; McCarty et al., 2022). Spikes are short-timed high-frequency content signals reflecting individual cellular activity. Neural synchronization can be evinced by temporally relating spiking activity to the background oscillations of LFPs (Fries, 2009, 2015), and this relationship has been documented across multiple brain regions and cognitive functions (Zarei et al., 2020). This spike-phase coupling has functional consequences, such that in rats the coupling between spike timing and the LFP phase in β oscillations represents sensory information, with an inverted bell-shaped tuning curve (Zarei et al., 2020).

LFPs are a highly effective means of exposing what state a given cortical region is in, since it captures general dynamics not specific to any individual cell, but also some cells that do not spike. In the same way that there is likely much information available at the LFP level that is not represented at the single-unit level (Kay and Lazzara, 2010)

– dynamics and emergent properties exist that can only be detected at the level of summed activity over millions of neurons – so too is it expected to be the case that there are aspects of cortical computation that are only represented at the interactional, global scale, and not at the LFP or spike level. This is a basic presupposition of ROSE, but also many other frameworks in cognitive neuroscience (e.g., Pessoa, 2023). An exemplar candidate process here is working memory, which seems to involve discontinuous bouts of spiking activity, as opposed to steady-state neural dynamics, with cortical rhythms being thought by some to control these dynamics, implementing executive control and allowing us to direct our memory and attentional resources at will (Buschman and Miller, 2022).

Relating R to O is far from trivial. Consider how communication-through-coherence has typically been assumed to reflect phase-synchronization between oscillators. Recent work suggests an alternative mechanism, through which coherence is the consequence of communication and emerges because spiking activity in a sending area causes post-synaptic potentials in the same but also other areas (Schneider et al., 2021). These authors identified afferent synaptic inputs rather than spiking entrainment as the principal determinant of coherence, opening up new directions for framing the relation between units and coherence. LFP coherence appears to be determined by two factors (Schneider et al., 2021): (i) coherence due to the direct contribution of afferent synaptic inputs; (ii) coherence between the sender LFP and the summed population spiking activity in the receiver. Coherence therefore depends on connectivity strength and oscillation power, and does not need purely oscillatory coupling or spike-phase locking in a receiver.

Further complexities arise here. Perhaps the biggest obstacle concerns the observation that cognitive processes do not necessarily lead to an increase in oscillatory power, and there is no necessary or clear connection between neural activity and computational complexity. For example, rhythms may emerge under stationary conditions and for low-dimensional sensory inputs. Schneider et al. (2021) found in monkeys that β coherence and power were most prominent during fixation, not during memory and movement. In macaque V1, high γ oscillations have been found for low-dimensional visual stimuli highly redundant across space, but these oscillations disappear for more salient stimuli (Uran et al., 2021). Genuine inter-areal phase-synchronization has been extensively documented, particularly in hippocampal θ, but the point to be made here is that although the basics of ROSE are well-supported, the specific mechanistic, causal interactions between levels remains very much an open question.

Nevertheless, successfully relating the two fundamental signals discussed in this section (spike-LFP) can provide us with a comprehensive explanation regarding the neurobiology of cognition (Perge et al., 2014). Since many signals picked up by the LFP will also very likely be able to be found at the unit level, care must be taken to map out assembly-level effects from single-unit responses and also from multiple, clustered single-unit responses, to establish which levels of linguistic resolution can confidently be mapped to units.

*Gain-Field Mechanisms*

I have so far presented a framework for representing and accessing basic features. However, there also exists a plausible neural code for representing multiple features at once, as in the case of feature-bundles that compose into objects manipulable and searchable by syntax, in the form of a gain-field mechanism (Botvinick and Watanabe, 2007). Gain-field mechanisms can conjoin information coded at the spike level across distinct domains, and in the case of lexical features, we might expect portions of posterior middle temporal cortex (serving as an interface between conceptual representations in anterior temporal cortex, and thematic representations in angular gyrus; Matchin and Hickok, 2020) to be tuned to the product of two variables (conceptual, event, etc.), and would hence exhibit a preference for a specific conceptual and thematic space. For instance, research utilizing fMRI suggests that the superior temporal sulcus is subdivided into regions that encode agents, verbs, and patients, independent of temporal order (Frankland and Greene, 2015). Pushing these ideas further, posterior middle temporal gyrus and inferior parietal lobe both respond to conceptual information from three separate input modalities (action, motion, sound), whereas other cortical areas responded in a bimodal or unimodal manner (Kuhnke et al., 2022), suggesting that single-unit responses in these two regions will likely code for conceptual features called upon by complex lexical instructions triggered by sensorimotor γ.

The posterior temporal sulcus is well-placed to organize much of this activity. It is positioned next to areas responsible for lexical search (posterior middle temporal gyrus), conceptual combination (anterior temporal lobe) and verbal/event semantics (angular gyrus). This argument is made with great force in Matchin and Hickok (2020). Compare this with the organization of inferior frontal cortex: sub-portions recruited for semantic control are "topographically sandwiched" (Chiou et al., 2022) between the multiple-demand and default-mode systems, and intracranial recordings can also evince distinct topographies for frontal control regions in high γ activity (Assem et al., 2023). This mosaic-like architecture of frontotemporal language networks nodes (Murphy et al., 2022a) can likely inform the complexity architecture of ROSE, such that cortical sites of clear functional tesselation are predicted to be the very same sites of close interactivity between the ROSE components.

What other neurobiological details are relevant here? It is known that neurons can participate in multiple large networks, simultaneously, via firing at difference frequencies (Bucher et al., 2006), contributing to all components of ROSE. Neurons in medial temporal and medial parietal cortices (Woolnough et al., 2020) would be readily capable of coding the product of multiple variables along category-sensitive cortices, given the categorial topology of these regions (see Murphy, 2020c for limitations of gain modulation models for higher-order aspects of language, beyond the lexical level). Single-unit recordings and cortical stimulation mapping in humans also point to medial temporal lobe as coding for aspects of subjective memory and conceptual recall (Fried, 2022), overlapping with cortex responsible for early orthographic lexical access. A common set of mechanisms for gain modulation have also been identified: GABAergic inhibition, synaptically driven fluctuations in membrane potential, and changes in cellular conductance (Ferguson and Cardin, 2020).

Although ROSE simply requires features of language-relevant representations to be coded at the spike and ensemble level, indexed by gain-field mechanisms, other recent insights into single-cell computation are emerging that might refine this framework (Gallistel, 2021). Although synaptic weight remains the focus of information models, neurons are eukaryotic cells with multiple information storage and processing mechanisms (Fitch, 2021), including aspects of wetware (protein phosphorylation, gene transcription) and cell morphology (dendritic form), and features of non-synaptic information-processing may contribute to the development of the R component of ROSE, in addition to R-to-O mappings. For instance, computation via wetware is substantially less expensive than cell signalling (Sterling and Laughlin, 2015). The single cell is ultimately where genetics, thermodynamics and theories of computation can meet (Niven, 2016). Neuronal excitability itself is an intracellular process interacting with synaptic plasticity, capable of being transferred via mRNA transplants across cells (Abraham et al., 2019; Lisman et al., 2018). If one is also concerned with issues of computational efficiency in syntax (Murphy et al., 2022b), intracellular computations (pertaining to synaptic weights, electrochemical dynamics, protein phosphorylation, gene transcription, and cell morphology) appear to operate near the noisy coding optimum (Riecke et al., 1997) and are also orders of magnitude less costly than inter-cellular signalling (Gallistel, 2017). A genuine explanation for basic properties of linguistic information access may be in the near future achievable via this reverse-engineering perspective of basic properties of cellular computation, in particular if we follow Prasada (2021) in assuming that Gallistel's (2021) observations about the single-cell basis of memory must also be applied equally to conceptual representations relevant to higher-order cognition. For instance, memories for abstract quantitative facts, concerning durations, distances, numerosities, rates and probabilities, are claimed by Gallistel (2021) to be stored in terms of representations that depend on information-bearing molecules within neurons. Basic human concepts manipulated by language also bear abstract and quantitative dimensions to them (e.g., concrete concepts and their individuation criteria via numeric quantification; Gotham, 2016) that exhibit considerable internal complexity (Chomsky, 2000), potentially rendering them compatible with this scale of neural representation.

As Trettenbrein (2016) summarises the relevant literature: "[M]emory persists despite synapses having been destroyed and synapses are turning over at very high rates even when nothing is being learned". Pursuing this intuition, Ryan et al. (2015) suggest that synapses should be understood as providing "access points" to information already stored within cells. Linked memories furthermore appear to share synaptic clusters within the dendrites of overlapping neural populations, and the locus of protein synthesis shapes the structure of the memory trace (Kastellakis et al. 2016). Kristan Jr. (2016, p. 950) adds another note of caution: "[S]ome perfectly good neurons have no processes, some vertebrate neurons do not generate action potentials, and some small (less than a millimeter in any dimension) invertebrate animals get along just fine without fast action potentials in any of their neurons."

How can we begin to relate these issues of basic information processing to other scales of complexity? Concurrently, many researchers in systems neuroscience are embracing Chuderski's (2016, p. 1) assessment that "[c]ross-frequency coupling may serve as the optimal level of description of neurocognitive processes, integrating their genetic, structural, neurochemical, and bioelectrical underlying factors with explanations in terms of cognitive operations driven by neuronal oscillations". It is in this spirit that ROSE aims to mechanistically link both single-unit and population-level dynamics for natural language syntax. This is the theme we will turn to beow.

*Cells to Circuits*

One final source of tension here, though certainly not unresolvable, is between Sherringtonian and Hopfieldian views of neural computation. This is essentially a distinction between anchoring computation around biophysical details and highly specified cellular connections (focusing on the transformation of signals by nodes in a point-to-point architecture, viewing cognition as the result of patterns of node-to-node connections whereby nodes transform representations), or dynamics of neural spaces and populations (focusing on representational spaces with computation considered to be the transformation between spaces; an extreme interpretation of the Hopfieldian view omits biophysical details altogether) (Barack and Krakauer, 2021). At the heart of ROSE (in particular, the transition from R to O) lies the question of which domains of language specifically demand attention via either of these views (the neuron doctrine vs. the population doctrine), and to what extent they can be unified. These concerns lead us into one of the most difficult challenges for modern cognitive neuroscience: varying across domains, brain areas, behaviors, and species, do single-unit responses possess explanatory power because they are part of a larger population, or, conversely, do population dynamics possess explanatory power because of the inherent computations performed by single-unit constituents? For example, evolutionarily older brain areas may be explained better via Sherringtonian accounts, but more recently evolved structured might not.

Quite independently of these exciting challenges, the aspects of R reviewed here allow us to go some way from coding distinct features to feature-bundles. Our next topic concerns operations over these objects.

## Operation

It should hopefully by now be clear that ROSE makes direct predictions for what kind of experimental paradigms will be optimally utilized across distinct recording scales, and this becomes highly salient with the case of basic linguistic operations (O). The current model will assume that high γ activity indexes operations pertaining to transformations of R. γ oscillations can be generated via the reciprocal interaction between excitatory glutamatergic and inhibitory GABAergic neurons (Welle and Contreras, 2017), although their comprehensive neurochemical profile remains unsettled. High γ is assumed to index local cortical activity (multi-unit activity) (Leszczyński et al., 2020). Adopting assumptions in Murphy (2020a), narrowband and broadband γ parcellate representations of distinct complexity during coupling with θ, such that broadband γ transforms complex linguistic features and narrowband γ extracts more elementary representations (for empirical support of this complexity contrast outside of language, see Saleem et al., 2017; Storchi et al., 2017). In this section, I will focus more on the basic properties of high γ activity, and will return to these concerns of low frequency interactivity for the S and E components.

*Semantics of γ*

A fundamental assumption of ROSE is that much (perhaps all) of the relevant space of computation at the O level takes place within the varying dimensions of γ. It is therefore necessary here to consider the different types of γ rhythms in cognition, rather than viewing γ as a monolithic unit. The integration between broadband and slow (30–50Hz) γ is an interesting case in this respect. Broadband γ correlates with neuronal spiking; spiking is phase-locked to slow γ; and, in turn, fast γ exhibits coupling with slow γ (Bahramisharif et al., 2016). Intracranial recordings suggest that, in contrast to fast γ, slow γ does not seem to be triggered in a strictly item-specific manner, and so slow γ power likely does not carry information via amplitude but is rather responsible for temporally organizing other signals (Bahramasharif et al., 2018); see also Bartoli et al. (2019). It is possible that the sites of slow γ rhythms act as relay stations between the sites of slow rhythms and the ultimate representational content accessed via broadband γ clusters (triggering the R level), necessitated and structured by specific projections. Indeed, this may demand an expansion of the δ-θ-γ code (see below) in Murphy (2020a) to a δ-θ-$γ^L$-$γ^H$ code (L = low, H = high).

Recent experimental work helps inform our understanding of the processes indexed by γ. Woolnough et al. (2021) show that broadband high γ indexes orthographic-to-lexical transformations. Other recent work documents γ signatures of basic lexical conversions during reading aloud single words (exception, regular, pseudowords) (Woolnough et al., 2022a): mid-fusiform cortex encodes lexicality; word frequency is encoded by mid-fusiform (earliest), inferior frontal and inferior parietal cortices; and orthographic neighborhood sensitivity in found in inferior parietal sulcus. This points to a distributed network for distinct features that compose into lexical items. Or, more specifically, it suggests that these regions are involved in regulating spiking activity across multiple sites encoding these features, potentially via spike-phase coupling or a similar mechanism. Other portions of cortex are likely dedicated to varying degrees to aspects of R that are accessed via O; for example, living and non-living nouns can be selectively impaired (Bi et al., 2016) and neurally represented (Chan et al., 2011), whilst nouns and verbs are also selectively impaired (Daniele et al., 1994) and neurally represented (Yang et al., 2017).

High γ is also a strong candidate for indexing the mapping of other basic psycholinguistic features, such as mappings between ortho-phonological information and meaning (de Zubicaray et al., 2022). Angular gyrus is a likely hub for γ-encoded O transformations for higher-order verbal semantic R units, being implicated in thematic/event structure (Graves et al., 2022) and the reconstruction of semantics from word-level phonology (Junker et al., 2023). Lastly, low frequency MEG activity in anterior temporal lobe and ventromedial prefrontal cortex seems to be sensitive to properties of conceptual combination (e.g., conceptual specificity), rather than syntactic structure itself (Pylkkänen, 2019), and future research could explore how these regions coordinate the activity of faster, cross-cortical activity likely recruited for accessing the relevant R features.

Consider next the results in Artoni et al. (2019). These authors constructed homophonous phrases with the same acoustic content but which are interpreted as either noun phrases or verb phrases depending of their syntactic context. Using intracranial recordings, they showed that VPs – semantically more complex than NPs – elicited greater activity in the high γ (150–300Hz) range than NPs across language-relevant areas. Again, we see clear involvement of high frequency activity in representational transformations (converting identical sensory information into semantically distinct data structures, with local cortical activity increases indexed by high frequency power modulations).

Other recent intracranial work directly supports this hypothesis concerning high γ indexing of lower to higher-order feature transformations. Focusing on auditory cortex and surrounding sites of superior temporal gyrus, Keshishian et al. (2023) showed a transition from the representation of early acoustic, phonetic and phonotactic features in high γ, to the later representation of lexico-semantic features, with these higher-order semantic features being represented further away from primary auditory cortex. The representation of earlier features was sustained throughout, with the transformation of speech sounds to lexical meaning being indexed by high γ.

Overall, these studies suggest that high γ is involved in the *conversion* or *mapping* of primitive units to larger feature-bundles that compose into complex linguistic objects. These linguistic representations often seem to be used by the same cortical sites in both comprehension and production (Hu et al., 2022). The cortex appears to implement a range of reusable computations (Marcus et al., 2014), permitting this generic mapping process to take scope over (evolutionarily) novel feature spaces. For example, prefrontal cortex may host circuits supporting variable binding, sequencing, gating and working memory storage (syntax-external demands), while lateral posterior temporal cortex might more reliably and efficiently subserve cross-modal semantic integration and supraordinate categorization (syntax-internal demands).

This theme of conversion and mapping is directly related to concerns of the type/token distinction. Other neural models of language have addressed the type/token issue comprehensively (Baggio, 2018), providing a level of depth that I will not attempt presently. I will simply note that, at least in terms of the semantic network for orthographic language comprehension, a model for types, tokens and referents has recently been proposed from a large cohort of intracranial recordings (Murphy et al., 2022a). This network implicates posterior superior temporal sulcus (pSTS) and inferior frontal sulcus as coding for types in high γ activity, with processes pertaining to lexical search and semantic coherence implicating a considerably widespread network (medial parietal cortex, parahippocampal cortex, ventromedial prefrontal cortex, posterior middle temporal gyrus). The role of medial parietal cortex here is particularly important in the context of discussions of endogeous attention, sensorimotor conjunctions, and prediction/error correction (Alexander et al., 2023). With respect to ROSE, it is notable that pSTS and closely neighboring portions of posterior temporal cortex code for *all* three components of meaning (type, token, referent). In many ways, the present assumptions about pSTS hosting types (i.e., types at the R level via sparse coding of single-cell features, or more redundant features further away from pSTS and towards neighboring temporal sites) and its interactions with frontal and parietal cortices generating tokens and referents provides support for core aspects of the MUC model (Baggio & Hagoort, 2011; Hagoort, 2013), but with the additional neurocomputational architure scaling from single-cell to ensemble to a coordinated, multiplexed network.

Lastly, are there any similar proposals outside of the language domain that fit the present model? A potential relation between the scales of complexity for R and O can be found in the domain of motor chunking. During motor sequence learning in monkeys, single neurons in the prefrontal cortex and basal ganglia begin by emitting a burst of spikes to each action in the sequence (an individual motor representation), but then shift to firing only at the first and last items, seemingly indexing chunk boundaries (Fujii and Graybiel, 2003; Jin et al., 2014). This moves us from single-cell representations of acquired motor habits, to the more dynamic operation of establishing a more complex representation.

## Structure

Acquiring and manipulating embedded tree structures, in a recursive fashion, is a defining human achievement (Fitch, 2014). Even basic structures exhibit sensitivity to syntactic identity (Lasnik, 2017; Martin et al., 2020). Sentence structure, in variable ways, guides a number of linguistic processes such as lexical recognition and semantic integration (Lowder et al., 2022; Pietroski, 2008). Discovering a feasible neural code for recursive syntactic composition would contribute to an explanation for the generativity of human intelligence and, potentially, the acquisition of composable functions more broadly (Dekker et al., 2022).

Thus far, I have mapped out candidate neural mechanisms for elementary linguistic representations and their manipulation. This section will begin by outlining the profile of supraordinate structures aimed to be captured by ROSE, before progressing to how this model converges with existing research.

*Basic Structures*

The syntactic component of language involves the construction of binary-branching hierarchically organized sets via the operation MERGE. This operation initiates with lexical access, manipulating objects from the lexicon or objects part of the syntactic workspace. Given the set {X, Y}, we can either select a new lexical object and MERGE it, to form {Z, {X, Y}}, or we can select an existing object to form {X, {X, Y}} or {Y, {X, Y}}. These sets are then labeled and given a syntactic identity, based on which element is most structurally prominent and easiest to search for (i.e., Z in the structure {Z, {X, Y}}). MERGE can also derive some set-theoretic properties of linguistic relations, such as *membership* and *term-of*, as well as the derived relation of *c-command* (sister of). In an attempt to render theoretical models more psycholinguistically compatible, MERGE has recently been defined as an operation on a workspace and its objects (Chomsky et al. 2019), formalized as follows (WS = workspace; P/Q = workspace objects; X = additional elements):

WS = {P,Q,…}

MERGE(P,Q,WS) = WS' = {{P,Q},$X_1$,…,$X_n$}

A basic case of traditional External MERGE (responsible for argument structure) can be represented as:

WS = [X, Y, Z] → MERGE(X, Y) → WS' = [{X, Y}, Z]

In contrast, we can also implement Internal MERGE ('movement'; responsible for discourse- and information-related properties):

$$WS = [\{X, Y\}, Z] \rightarrow MERGE(X, \{X, Y\}) \rightarrow WS' = [\{X, \{X, Y\}\}, Z]$$

The execution of MERGE relates ultimately to structural relations, rather than to independent pressures like communicative priorities (Murphy, 2020b). As Hagoort (2023) comprehensively reviews, the computational system of language seems to be related to neural signals integrating perception and action, providing humans with novel modes of planning and interpretation, whereby lexical units and unification processes like MERGE provide "an imaginary space that transcends the influence of direct perception-action cycles" (Hagoort, 2023, p. 2). This positions MERGE outside of the direct influence of sensorimotor γ signatures, for example.

I will be concerned here with exploring candidate neural mechanisms for MERGE in an effort to construct the S component of ROSE, putting aside some of the performance-related issues pertaining to real-time parsing, which require deeper empirical exploration.

*Syntactic Signatures*

I will assume here that the processing correlates of MERGE and labeling take place after posterior temporal δ phase-locks to θ either in neighboring temporal sites or inferior frontal sites, depending on the category of structure being composed. Dorsal arcuate fasciculus, connecting posterior temporal with posterior inferior frontal sites, contributes to syntactic encoding here (Goucha et al., 2017), whereas more ventral portions, and also more anterior temporal regions, are recruited for conceptual-semantic processes (Pylkkänen, 2019). The importance of posterior temporal cortex for the basic property of syntactic computation cannot be emphasized enough (Matchin and Hickok, 2020). For example, the native language of polyglots is processed locally within posterior temporal cortex, with only minimal frontal involvement, and frontal cortex is recruited more for their many non-native language representations, presumably demanding more plastic, controlled representational access to variations in functional morphosyntactic structure (Malik-Moraleda et al., 2023). Lesion-symptom mapping of language impairments in individuals with brain tumors also suggests that while single-word production deficits are associated with inferior parietal cortex, phrase and sentence production deficits are additionally associated with posterior temporal gyri (Ntemou et al., 2023).

A range of converging evidence for the role of neural oscillations in posterior lateral temporal cortex coding for natural language structure is emerging (e.g., Matar et al., 2021). Importantly, the role of endogenous oscillations in neural computation, over and above the role of entrainment, has been recently documented (Duecker et al., 2021; van Bree et al., 2021). The δ-θ frontotemporal interactions in S are assumed to influence medial temporal and frontal θ-γ coupling pertaining to representational bundling. This model is also supported by the low frequency tracking literature, which finds stronger δ signatures for sentence-tracking, and also chunk-tracking, in temporal cortex over frontal cortex (Jin et al., 2020). By focusing on syntactic hierarchies, I put aside here issues of morphophonology, or morphosyntactic linearization, which seem to implicate inferior frontal cortex (Matchin and Hickok, 2020; Sahin et al., 2009), and which may also recruit aspects of the θ-γ code for linear sequencing (Heusser et al., 2016; Lizarazu et al., 2023).

Neighboring domains of cognitive neuroscience provide some converging insights here. In a reward-based decision-making task utilizing EEG, Riddle et al. (2022) found that reward-evaluation was marked by frontal θ phase coupled with parietal γ, whereas goal-directed behavior was positively correlated with coupling between frontal δ to motor β amplitude. The executive, supraordinate signatures of δ-driven coupling seem clear.

In the language domain, Keitel et al. (2018) analysed speech tracking in source-localised MEG data and found the following structures were tracked at particular bands: phrases (0.6–1.3 Hz), words (1.8–3 Hz), syllables (2.8–4.8 Hz) and phonemes (8–12.4 Hz). The δ rhythm which tracked phrases was coupled with β power in motor areas, which likely implements top-down speech predictions. Across humans, monkeys and mice, β seems to be generated by nearly synchronous bursts of excitatory synaptic drive targeting proximal and distal dendrites of pyramidal neurons (Sherman et al., 2016).

Keitel et al.'s (2017) MEG study of auditory cortical δ entrainment, and its interactions with frontoparietal networks, provide further insights. During intelligible speech processing, δ entrained with three networks: δ-β coupling occurred between δ in the left anterior superior temporal gyrus and β in left-lateralised medial orbitofrontal areas, which the authors claim reflected predictive top-down modulations of auditory encoding. δ-α coupling occurred between δ in left Heschl's gyrus and α in anterior superior temporal gyrus. Finally, δ in right posterior superior temporal gyrus coupled with predominantly right-lateralised parietal θ, which likely reflected semantic memory engagement. Auditory δ entrainment thereby seems to be influenced by left orbitofrontal β and parietal θ. These δ-θ interactions may index the mapping of acoustic-phonemic processing to more specific morphosyntactic features, which are then subject to coordination by other δ-driven mechanisms. Relatedly, using MEG Klimovich-Gray et al. (2021) showed that the joint support of temporal and contextual predictability reduces word and phrase-level cortical tracking in δ, likely due to less parsing effort being needed in these contexts, supporting the higher-order inferential function of δ.

Nevertheless, some recent research has pointed to potential concerns with this frequency-tagging paradigm when used with natural speech, such that models invoking endogenous synchronicity may be better placed to explain the spatiotemporal dynamics of syntax (as opposed to purely chunking via exogenous entrainment). The paradigm used by Keitel and colleagues has technical issues that may prevent it from exposing phrase-specific responses: Zhang et al. (2023) show via simulations that the 1 Hz phrase-level tracking largely originates from the pauses in natural speech. The dynamics that arise through faster δ-driven and θ-driven interactions seem less susceptible to this problem.

I have focused here on concrete neural mechanisms, mostly abstracted away from parsing details. Questions of timing are here purely empirical, but interactions between these layers of the ROSE processing hierarchy afford a considerable degree of rapid flexibility. Consider how hierarchical phrase composition appears to be indexed around 200–300ms after the composition-permitting item (Hale et al., 2022; Murphy et al., 2022e). Just because δ can host, for example, three cycles per second (3 Hz), this does not mean that, therefore, syntactic computation must occur over relatively long spans (~300–350ms). The δ-coordinating activity takes place mostly in the period immediately after the trough of the cycle. Furthermore, only a portion of this post-trough period will be directly responsible for phase-amplitude coupling dynamics, not the entire trough/peak cycle. As such, this will take place over much faster periods, likely within ~80–165ms chunks, or faster (keeping purely with 3 Hz as an example). These windows are compatible with MEG-derived and EEG-derived time courses for syntactic binding and semantic composition (Murphy et al., 2022e; Pylkkänen, 2019), with effects lasting around ~90–120 ms for conceptual composition, argument structure encoding, and sensitivity to long-distance dependencies.

In addition, recent work has begun to push back against the traditional, Friederician 'syntax-first' processing model of sentence comprehension, and has shown that syntactic categories do not appear to be identified first and also do not condition lexical-semantic integration, demanding a revision to strictly seriel models of phrase structure comprehension. For instance, instead of observing a LAN-P600 complex for syntactic category violations, or only a P600 effect, Fromont et al. (2020) systematically observed an N400-P600 effect. In terms of timing relevant to ROSE, this work indicates that the initial semantics-driven properties of R and O can readily integrate with each other over time windows partially overlapping with slower (and low frequency) signals integrating syntactic information.

Reinforcing the importance of the generation of supraordinate categories, in MEG decoding of speech for basic adjective-noun phrases, "the head of the phrase, the noun, engages a much more stable set of neural representations than its modifier, the adjective" (Honari-Jahromi et al., 2021), potentially due to its labeled status. We also find β oscillations in posterior temporal cortex exhibiting sensitivity to syntactic complexity (Matar et al., 2021), positioning this as a core component of identity-attribution in syntactic labeling (Murphy, 2020a), responsible for aspects of maintenance as opposed to active structure-generation (δ-θ).

*Mapping from O to S*

One of the earliest stages of transitioning from O to S likely comes in the form of activating in parallel multiple (and potentially overlapping) unit groups coding for R but being manipulated, mapped, or transformed by γ. It has recently been discovered via human intracerebral recordings that high γ oscillations (70–200Hz) can exhibit long-range phase synchronization, which appears behaviorally relevant in a response-inhibition task (Arnulfo et al., 2020). This broadens the scope for simultaneously active (bound) linguistic features to be organized into a set via cross-cortical manipulation (coordinated, under ROSE, via low frequency activity in the service of workspace construction; see below). This approximates the formal notion of unordered binary set-formation (Chomsky et al., 2019) more so than the linear feature-clocking of phase-amplitude coupling suggested in Murphy (2020a). The joint coordination and activation of cross-cortical linguistic features (free from any serialization constrained via phase-amplitude coupling) may constitute one of the earliest stages of phrase structure building, prior to morphosyntactic linearization.

*Low Frequency Structure-Tracking*

One of the most important research directions in neurolinguistics that pertains specifically to the S component of ROSE is the low frequency tracking literature (Di Liberto et al., 2018; Ding et al., 2016; Jin et al., 2020; Keitel et al., 2018; Lu et al., 2022a, b; Makov et al., 2017, Schmidt et al., 2022). The most recent advances here have shown separable low frequency repsonses to syntactic rule-based chunking independently of lexico-semantic features (Lu et al., 2022a, b), constituting an independent neural process from basic statistical analysis of word features. Other seminal work here includes Kaufeld et al.'s (2020) demonstration of how δ-band scalp EEG activity is modulated by lexically driven combinatorial processing beyond prosody. Low frequency oscillations can appear in synchrony with chunks of structure, varying also with flexible time scales of speech presentation.

What makes low frequency behavior highly relevant here is that certain tracking effects may be unique to higher cognition. Multi-syllabic zebra finch songs do not result in slow neural activity matching the time scales of specific chunks, and we instead find bursts of discrete, transient activity for each syllable, rendering this appropriate for item-based state transition models, but not models invoking syntactic structure (Long et al., 2010; for further discussion, see Schlenker et al., 2022). There is also recent evidence for the role of a category-sensitive phrase-building (labeling) mechanism driving the cortical tracking (via low frequencies) of hierarchical linguistic structures (Burroughs et al., 2021), as predicted in Murphy (2020a) and which is in line with the present ROSE architecture.

Despite the above reservations about frequency-tagging paradigms when applied to naturalistic speech, neural envelope tracking seems vital for a number of comprehension mechanisms, and is decreased in individuals with post-stroke aphasia (De Clercq et al., 2022). Low frequency cortical activity tracks both overt and covert prosodic changes, and this tracking interplays with syntactic processing. Using scalp EEG, Glushko et al. (2022) showed that alignment of syntax and prosody boosted EEG responses, whereas their misalignment had the opposite effect. This was true for both overt and imagined prosody. A range of studies have shown how prosody serves as an aid to syntactic parsing, such that speakers tend to use stronger prosodic boundaries to mark left dependencies (Degano et al., 2022). Yet, as Glushko et al. (2022) demonstrated, some signature of syntax is still provided with δ tracking, even if constituent structure is not uniquely encoded via δ frequency-tagging.

Consider also the findings of Rimmele et al. (2022). Using MEG, these authors had participants listen to disyllabic words presented at a rate of 4 syllables per second. They found that lexical content (as opposed to syllable-transition information) activated a left-lateralized frontal and superior/middle temporal network. The authors discovered δ-θ phase-amplitude coupling in middle temporal gyrus and superior temporal gyrus in conditions with lexical, transitional and syllable information, but not with only syllable information, suggesting that syllable information is exploited for lexical-level processing. As noted above, ROSE places δ-θ interactions as part of a more general syntactic, categorization process, not unique to 'word'-level elements, applying also to sets of lexical objects to form phrases. Purely at the level of speech properties, there appears to be a role for θ tracking throughout sustained acoustic fluctuations, but a greater involvement of δ tracking at speech onset (Chalas et al., 2023).

Other supporting evidence for ROSE comes from Gross et al. (2013), who discovered δ-θ phase-phase coupling during speech perception, with θ also modulating the amplitude of γ, suggesting that δ phase likely modulates θ in distinct ways depending on the computation. Halgren et al. (2017) also revealed that in the human brain cortical δ phase "robustly modulates theta power, with an increase in theta-band power during the falling phase of the ongoing delta rhythm". It is currently an open question what type of parsing procedures are indexed by various types of δ-θ relations (e.g., phase-phase versus phase-amplitude relations).

*Dissociating Structure and Meaning*

Contrasting with low frequency signatures of structure, high γ appears to encode elements of meaning at the minimal phrase level (Murphy et al., 2022e) and, incrementally, in orthographic sentence processing (Woolnough et al., 2023) – but not structural identity at the complex multi-phrase, syntactic node-tracking level (Woolnough et al., 2023). Importantly, Woolnough et al. (2023) failed to detect γ signatures of syntactic node structure, as initially reported in Nelson et al. (2017), likely due to Nelson et al. not including lexical frequency in their model; once this is included, structure-related parsing effects disappear in high γ. Importantly, Woolnough et al. (2023) found increased γ for Jabberwocky over Pseudoword lists, but this was crucially punctuated and not sustained across multiple words in the sentence. As such, this likely represents a signature of the basic structure-building operation, MERGE, and these Jabberwocky γ effects also lasted around 200–300ms, seemingly the right resolution for the timing of syntactic composition (Pylkkänen, 2019). When considered alongside the failure to find γ signatures for structure-building parsing models, these results conspire to suggests that operations (O) can be detected transiently via γ, and that semantic information is represented here more robustly, while structural identity itself (S) is represented more globally in low frequencies.

While these results are in line with the framing of conceptual composition in ROSE, how can we address this apparent mismatch between high γ sensitivity to phrase and sentence meaning, but not sentence structure? One thing to consider is that constructs such as the labeling algorithm from theoretical syntax may not easily map to psycholinguistic factors (Murphy, 2023a), and so it seems that multi-unit syntactic node opening/closure might also not be encoded in high γ. Although the signal from a basic, minimal compositional scheme might be strong enough to be detected, driven likely by its semantic properties unique to the phrasal over lexical configuration, upon encountering more diffuse representations across multiple phrases, this may become difficult at the purely structural level. Hence, in terms of basic semantic composition operations, and perhaps also transient instances of MERGE, high γ seems sensitive. Indeed, propositional sentence-level meaning is represented in fMRI as a distributed network, not specific to any specific site (Anderson et al., 2021), already pointing to the functional utility of global mechanisms like low frequency coupling and phase synchronization. Other converging evidence for the present model comes from Toneva et al. (2022), who show that inferred sentence meaning ('supra-lexical' meaning) can be detected via fMRI in the posterior superior temporal sulcus but *not* via MEG, suggesting again that local cortical computation via synchronized pyramidal cell firing does not code for supraordinate elements of complex syntactic structure building.

As such, structures and storage may need separate scales of encoding, which under ROSE come in the form of low frequency synchronization and cross-frequency coupling. Operations (O) can occur at the intra-lexical or extra-lexical level, depending on the type of featural manipulation being carried out: transforming graphemes to phonemes, or to conceptual representations, or triggering a cluster of R features that go beyond the simple summation of discrete features, or building incremental sentence meaning or a situation model that is independent of structural format.

In effect, single cells under ROSE can represent formal and semantic features (similar to how they appear to represent various sound features; Leonard et al., 2022) such as root elements and categorization elements that assemble into basic syntactic objects, but structural information may be a Hopfieldian concern of population dynamics, calling upon more global (cross-cortical) and inter-areal coordination (as opposed to category-specific, unit-specific memory storage).

Quite suitably for this purpose, cross-frequency coupling has been widely linked with information integration across neuronal populations. Bybee et al. (2022) propose a role for cross-frequency coupling in the memory of discretized phase patterns, based on a principle in analog computing (subharmonic injection locking). Using a novel coupled oscillator Q-state phasor associative memory (PAM), Bybee et al. (2022) show that the presence of coupling increases the memory capacity in plastic synapses. If the present ROSE architecture is appropriate for language, then specific, memorized syntactic configurations (akin to Hagoort's syntactic schemas in posterior temporal cortex, or Matchin and Hickok's treelets) may be stored via types of discretized phase patterns exposed by Bybee et al. (2022), readily and efficiently called upon via cross-frequency coupling. With respect to related neural mechanisms, silent synapses (Vardalaki et al., 2022) in lateral posterior temporal cortex may be called upon ('unsilenced') to store new complex syntactic schemas for rapid recall, both during learning and short-term consolidation, coordinated by the low frequency ROSE code.

*Mapping from S to E*

We have so far presented candidate neural mechanisms for R, O, and S, and have also outlined means to map R to O, and O to S. Are there other potential means to further link R, O and S, and is there a means to further map S to E? This section will discuss vector codes and traveling oscillations as a means of positively answering these questions.

The present system of low frequency modulations of faster local cortical activity can serve to implement the algorithmic vector symbolic architecture of Eliasmith et al. (2012). Vector codes take scope over a fixed population of neurons and determine approximate values of activity across this cohort (e.g., active vs. silent). As mentioned, the existence of spike-phase coupling, seen as part of the more global system of low frequency phase coherence under ROSE, would permit a direct path from global low frequency synchrony, to a specific γ activation profile (across the trough-peak of the lower rhythm), to the vector code of this fixed population of neurons. It is known that spike-phase coupling increases with memory demand (Hanslmayr et al., 2021), but we currently lack any such evidence for syntactic composition demands, although this is predicted by ROSE. Even though this vector code itself remains obscure with respect to syntax (and requires a joint combination of single-unit recordings and LFP signatures), in principle ROSE can readily accommodate the more global dynamics that emerge from it. Indeed, ROSE leads to the prediction that syntactic node boundaries and complex, successive syntactic operations will have no robust vectorial representation at the single-unit level. For now, we can surmise that the monotonically increasing level of γ activity over the course of sentence processing (Woolnough et al., 2023) likely reflects this output of global low frequency coordination over this space of local vector codes.

One final mechanism ties together all presently discussed components of ROSE: traveling waves. Zabeh et al. (2022) show how traveling waves (in β) regulate neuronal spiking activity across space and time, being related to reward history in the monkey brain. These authors show how frontoparietal LFPs form traveling waves, and claim that wave direction is a proxy of intra-regional communication. This provides a feasible mechanism not simply for regulating δ-θ interactional dynamics in language, but also the dynamics of R units, controlling the co-activation probability of neurons in topographic networks.

Other recent work suggests that, at least in ventral temporal cortex, the propagation of event-related phase synchronization does not seem category-specific (i.e., there are few difference between faces and words) (Woolnough et al., 2022b), suggesting that traveling waves serve a more domain-general information transfer function in this portion of cortex, while the R and O scales serve domain-specific representational and computational functions. Note here that in order for this type of research to more stringently conclude that portions of cortex are category-sensitive or selective, we would have to detect the relevant activity profiles in the *absence* of stimuli triggering a given representation, not simply when subjects are shown pictures or words depicting these representations. The propagation of phase modulation "may represent the dynamic coordination of neural processing across spatially disparate regions […] modulating excitability of local neural populations" (Woolnough et al., 2022b). These results support a phase reset model of event-related potential generation, such that a resetting of the instantaneous phase of ongoing oscillations results in the ERP through creating a coherent superposition of phase components (Iemi et al., 2019).

In contrast, in association cortex distinct types of traveling waves (feedback, top-down) that are more category-specific seem to exist (Murphy, 2020a), which may encode higher-order information. Nevertheless, Bhattacharya et al. (2022), in their study of monkey lateral prefrontal cortex, did not observe any changes in traveling wave profile across different working memory task sample items, which is again consistent with the view that traveling waves have a 'meta' network function that is independent of item identity. Essentially, S codes specifically for recursive structure generation, and traveling waves migrate these representations over to relevant workspaces in E (see below).

Much other work suggests that traveling waves can create timing relationships fostering spike-timing-dependent plasticity and memory encoding (Sreekumar et al., 2020), and that they contribute information about the recent history of activation of local networks (Muller et al., 2018). Connecting the distinct levels of ROSE, microelectrode recordings in the human brain suggest that it also seems feasible for macro-scale waves to co-occur with micro-scale waves, which can be temporally locked to single-unit spiking (Sreekumar et al., 2020). Temporal coordination between traveling waves at different spatial scales, and between waves and spiking activity, points to a role for traveling waves in neural communication. The different types of

macro- vs. micro-scale traveling waves may form their own internal hierarchy of sensitivity to distinct structures or higher-order relations in language processing, e.g., global situation model maintenance vs. local phrase structure coordination.

A more ostentatious series of speculations concerns the role of brain shape in cross-cortical information transfer, such that the human brain has assumed a more globular shape than our closest relatives (Benítez-Burraco and Murphy, 2019), and this may contribute causally to explaining efficient cross-cortical synchronization, opening up new paths for traveling waves and inter-areal connectivity. Recent work in this connection has pointed to various geometric constraints on human brain function (Pang et al., 2023) and highlighted strong relations between functional connectivity profiles and critical episodes of neural regularity and complexity (Krohn et al., 2023), although presently little else can be said of direct relevance to ROSE.

For now, traveling waves seem suited to memory-related and encoding-related processing, dynamically coordinating the local storage of cortical information, forming the foundation of any notion of syntactic workspace arising out of implementations of ROSE. Bhattacharya et al. (2022) document rotating traveling waves relevant for working memory in prefrontal cortex in the monkey brain. These ranged from $\theta$ to $\beta$, and during task performance some waves (mostly $\beta$) increased in number in a given direction greater than during baseline conditions. Soroka and Idiart (2021) propose a model for working memories under which $\delta$, $\theta$ and $\alpha$ are traveling waves, whilst simulatenously undergoing phase-amplitude coupling relations coding for the ordering of working memory items ($\gamma$) – and constraints on working memory size – by slower rhythms. In their model, the interactions between $\theta$ and $\gamma$ determine the allocation of multiple items, while interference between $\theta$ and $\alpha$ disrupts the maintenance of the current workspace. Under ROSE, a similar architecture exists but with the additional layer of hierarchical, executive coordination by $\delta$ for syntactic inferences over workspaces.

*Summary*

In brief, ROSE assumes that the amplitude of $\theta$ (involved in $\theta$-$\gamma$ bundling of elementary features for syntactic manipulation) is in turn coupled with the phase of structure-indexing $\delta$, with $\delta$-$\theta$ interactions alongside $\delta$-$\beta$ interactions providing an additional layer of phase coding to yield recursive structure formation necessary for natural language syntax (see Murphy, 2020a, for further cartographic details concerning specific frontotemporal network nodes). Notice that the neural codes outlined for structural inferences (S) and representational binding (R, O) are mechanistically related, but they are distinct in ways that lead to empirical predictions for the processing of logical and grammatical relations, such as negation, pronominal binding, agreement, and sentence coordination. These processes are predicted to yield high γ signatures over sites of representational binding (*à la* the binding of visual stimuli into faces or scenes; Woolnough et al., 2020; see also Yu & Lau, 2023), since while they are constrained by structural factors the ultimate signature is expected to be concerned with accessing a specific representational format (R) via sensorimotor mappings (O), even if this is to be guided by the more global, supraordinate labeled identity of the current workspace (S, E). Intracranial recordings of morphosyntactic agreement, for example, could probe this issue further.

**Encoding**

Syntactic comprehension seems to require multiple workspaces. For Chomsky (2023, Forthcoming), there are parallel search procedures for labeling. For Adger (2017), one workspace is used to construct syntactic phrases, and the other is used to temporarily maintain these units once they have been transferred. This topic constitutes the final component of ROSE: encoding structures after their generation (E). Following Murphy (2020a), this encoding aspect can be addressed as follows: the initial θ-γ code constitutes the first workspace (sensitive to lexical memory), and the subsequent δ-θ code constitutes the second (sensitive to supraordinate, hierarchical memory).

To illustrate, consider the sentence 'old men walk slowly'. During the comprehension of the first two words, the δ-γ combinatorial code coordinates the feature-bundling of the atomic data structures hosted by 'old' and 'men' (Brennan and Martin, 2020). At a minimum, this involves pSTS low frequency activity coupled with neighboring posterior temporal cortex but also cross-cortical sites responsible for the specific feature types in question (e.g., ATL, IFG) (Murphy et al., 2022e). θ-γ coupling maintains in short-term memory the relevant units (R) via high frequency activity (O) in a linear sequence (Heusser et al., 2016). At the transition between 'men' and 'walk', the supraordinate δ-θ code maintains the categorial (i.e., structural) identity of the object, in this case the negotiation between a multi-unit noun phrase and a more complex verb phrase hosting 'old men'. During the same period, the initial lexical memory code increases its number of θ-nested chunks due to the occurrence of 'walk'. The same transition occurs from 'walk' and 'slowly', with the exception that while the lexical memory code still increases in coupling strength, the hierarchical memory code would *decrease* closer to (but not identical to) its pre-verb baseline, due to the adjunction relation not demanding a revision of the hierarchical memory representation. The transfer of relevant lexical information (i.e., for categorization and labeling/search) would take place via interactions between these two neural codes, potentially via θ-θ phase-phase coupling or phase-locking of the θ-γ workspace and the δ-θ workspace. θ-driven dynamics effectively constitute the 'hand-off' of information after lexicality has been established by lower-level R and O processes, transitioning from encoding lexical memory to multi-object memory. Direct testing of these dynamics specifically with respect to syntactic workspace construction has currently not been undertaken, although much work has been carried out demonstrating increased θ-γ coupling in human hippocampus during memory formation (Lega et al., 2016), as well as enhanced frontal θ to posterior γ coupling (Friese et al., 2013).

Recent research is compatible with this model: Becker and Hervais-Adelman (2023) found that left posterior middle temporal gyrus θ-entrainment in MEG is associated with syllabic and lexical recognition during speech processing, while δ-entrainment across broader portions of posterior temporal cortex is associated with phrasal recognition. In a word and face processing task using MEG and intracranial recordings, Fellner et al. (2019) found that across a number of brain areas increases in γ power and decreases in θ power predicted memory formation irrespective of material, and that γ increases occurred significantly earlier compared to θ decreases, aligning with the general transition of γ (R/O) to θ and δ (S/E) complexes that conclude with the encoding component of ROSE. This is compatible with the Information via Desynchronization Hypothesis, which states that neural desynchronization increases the amount of information that can be neurally encoded since it allows discrete cells to transmit individual messages, as opposed to a group attempting to transmit the same message (Schneidman et al., 2011).

At the highest level of abstraction, we can summarize that δ-driven coupling constitutes the encoding of top-down syntactic information, while θ-γ coupling represents bottom-up (feature-to-assembly) level information driven more directly by perceptual inference. The remainder of this section will elaborate on this proposal and explore further connections with the above framework for R, O and S.

*Syntactic Memory as Phase-Synchronization over Successive Cycles*

Given that we are dealing here with domain-specific cognitive properties (as they pertain to linguistic structure building), are there any reasons to highlight the ROSE components as being suitably distinct from a neurocomputational perspective? Consider how there appears to be a broader range of cross-frequency coupling relations in the human brain relative to the anthropoid apes, with the human cortex exhibiting a species-specific level of richness in its cross-frequency coupling profile (Maris et al., 2016), pointing to interactional dynamics as being a possible major index of human-specific computational capacities. These assumptions are bolstered by recent work showing that individual differences in syntactic processing performance is related to white matter connectivity differences between nodes of the language network (Sánchez et al., 2022). These low frequency interactional dynamics seem to be encoded via Erdős-Rényi networks (Murphy, 2020a). Future modeling of oscillatory networks for language and their connectivity profiles (Thiebaut de Schotten and Forkel, 2022) crucially need to be centred around biophysical models that are inherently compatible with recorded neural signals, rather than intuitive models of oscillations that may lead to spurious detection of rhythms (Doelling and Assaneo, 2021; van Bree et al., 2022).

This species-specific richness in cross-frequency interactions may be principally related to the expansion-fractionation-specialization (EFS) hypothesis (DiNicola and Buckner, 2021). Evolutionary expansion of human association cortex may have allowed for an archetype distributed network to fractionate into multiple specialized networks (DiNicola and Buckner place special emphasis on frontotemporal areas implicated in higher-order language). One such network seems to be anchored around a cortical mosaic for linguistic structure and meaning along the pSTS (Murphy et al., 2022e), interacting with inferior frontal cortices during more complex functional grammatical structure-building (Woolnough et al., 2023), and under ROSE would form the basis of low frequency hierarchical coupling in the δ-θ range.

At the algorithmic level, the present conception of E is compatible with the framing of syntactic working memory as arising from continuous or discrete attractor states, implemented neurally via activity in recurrent networks (Wang, 2013) and the coordination of firing patterns via low frequency phase coding.

*Symbolic Computation*

Recent directions in the neurobiology of navigation and memory are relevant to the present conception of E. As reviewed in Kurth-Nelson et al. (2023), hippocampal-cortical sequence replay and encoding is not constrained to simply repeat past experience. Rather, this process is informed by an internal model of the world, generating representations of inferred entities not necessarily encountered physically. This active, generative capacity motivates the authors to propose that replay in the brain instantiates a form of compositional computation. A given replay sequence constitutes a set of entities strung together into a compound, whereby each entity is bound to a representation of its compound role, determining its function as part of a whole. This establishes a clear separation, with respect to composability, between entity and role (or semantics and syntax). While roles encoded by hippocampal-cortical interactions can certainly be spatial, they can also be non-spatial, and even non-spatial and non-Euclidean (Kurth-Nelson et al., 2023; Liu et al., 2019), potentially involving arbitrary roles such as 'verb'. The entity-role bindings currently explored empirically in humans are limited to things like 'which position' and 'which sequence', but if other roles like 'if', 'then' and 'else' can be encoded in a similar way then replay may form a viable candidate for a neurophysiological mechanism implementing symbolic computation. This compositional nature of replay is implemented via θ-γ (and also fast ripple) hippocampal-cortical interactions (Kurth-Nelson et al., 2023), mirroring closely the present assumptions of lexical feature sequencing and basic semantic compositionality being implemented via the same dynamics and high γ activity.

*Memory Transfer*

Other neural mechanisms are beginning to emerge for memory transfer that may bear some relevance for transferring chunks of structure to distinct cognitive systems. Interhemispheric transfer of working memories has been found to be achieved via left and right prefrontal cortex via θ synchrony, with transferred memory traces activating different ensembles than feedforward-induced traces (Brincat et al., 2021). Cortical feedforward processing has been shown to be mediated by θ and γ (Bastos et al., 2015). The assumption of ROSE is not simply that γ increases during maintenance and decreases during read-out, as in current working memory models (Lundqvist et al., 2018), but rather that this process naturally requires coordination over a set of distinct memory representations that are concurrently clustered via more global, supraordinate control mechanisms, which take the form of phase-amplitude coupling in δ-θ for syntactic category and hierarchical memory and θ-γ for 'lexical' (feature-bundle) memory (Figure 1).

As briefly mentioned, β increases seem to signal the build-up of syntactic predictions (Murphy, 2020a), but it also seems that β suppression in prefrontal cortex marks cortical disinhibition permitting working memories to be expressed. β suppression likely co-occurs alongside the encoding steps in ROSE (Figure 1), in keeping with current models, while β increases at different periods of sentence processing seem to index prediction (Murphy et al., 2022e), potentially in neighboring inferior frontal sites. The transfer of working memories from one portion of cortex to another occurs at around 120ms (Brincat et al., 2021), such that working memory representations decrease after the initial increase at the receiver, signalling a 'soft handoff' of information whereby some trace still remains at the initial construction site during this brief period. It may be that the duration of this handoff period for natural language (as opposed to the working memory tasks in the monkey brain probed in Brincat et al., 2021) increases with the complexity of the multiplexed, compressed signal of feature-bundles to be spell-out. Indeed, it is thought that interhemispheric communication is disrupted in certain dyslexias (Dhar et al., 2010). In addition, familial sinistrality has been found to impact the brain's responses to morphosyntactic violations (Leckey et al., 2023); the field currently lacks a consensus on the interhemispheric organization of syntax-specific and extra-syntactic comprehension mechanisms, and inter-areal analyses centred on oscillatory dynamics may well contribute here.

*Alpha/Beta Dynamics*

So far we have focused mostly on δ-θ signatures, but as indicated briefly above, other low frequency signatures seem to play essential roles in the regulation of form-meaning mappings and the integration of information from syntax-semantics into domain-general systems.

Keeping for the moment with syntactic memory, results reported in Gehrig et al. (2019) support a role for β in syntactic identity. These authors investigated speech memory representations using intracranial recordings in the left perisylvian cortex during delayed sentence reproduction in patients undergoing awake tumor surgery. Based on the memory performance of patients, they found that the phase of frontotemporal β represents sentence identity in working memory. The notion of sentential identity presupposes a labeled structure (i.e., CP, VP), seemingly represented (at least partially) by frontotemporal β. Converging with other literature (reviewed in Murphy, 2020a), β may represent aspects of the global cognitive set going beyond syntax-specific information to include conceptual and statistical information pertaining to sentence identity that is immediately captured by structure-building, whereas the δ-θ-γ interactions discussed above code for more specific language-internal compositional processes.

Intracranial recordings of auditory language comprehension implicate frontal α/β power (8–30 Hz) in phrasal prediction/anticipation (Murphy et al., 2022e). Pefkou et al. (2017) found that both θ and γ are sensitive to syllable rate, but only β power is modulated by comprehension rates and is insensitive to syllabic structure. This suggests, again, that θ power plays a more bottom-up role in feature-set construction (Figure 1) while β appears to be involved in the tracking or prediction of semantic or phrasal identity. Independently, θ power can code for lower-level stimulus features, but when coupled with δ, a more higher-order role can be established, likely depending on cortical site.

In parietal cortex, α enhancement seems to index syntactic working memory demands (Meyer et al., 2013). Much like the regulation of γ by α in control and attention mechanisms (Bonnefond & Jensen, 2015; Jensen et al., 2014), parietal cortex may play an important role in memory and complexity, such that low frequency rhythms originating in lateral or medial parietal cortex regulate the activity of γ-encoded O in lateral temporal cortex, and single-unit or assembly-encoded R in medial temporal or inferior frontal cortex.

Another likely site of syntactic working memory for E is inferior frontal sulcus, since this is implicated in task-related evaluation of phrasal meaning, and exhibits functional coupling with posterior temporal sites during structure composition (Murphy et al., 2022e), and is engaged for a number of semantic integration demands (Murphy et al., 2022a). As such, it is expected that lower frequency components in inferior frontal sulcus and inferior parietal cortex are functionally coupled with other low frequency components coding for S, discussed above, providing the read-out of structure into short-term encoding (Makuuchi and Friederici, 2013).

Turning lastly to the issue of memory training, using MEG Wang et al. (2021) trained native German speakers in syntactically complex German sentences via comprehension tests over a series of four days. They found that successful training strengthened the use of the dorsal processing stream, with working memory-related regions in inferior frontal sulcus demonstrating decreased γ (55–95 Hz) over the course of the four days. Future work testing ROSE could explore the electrophysiological dynamics of this region, likely a source of the δ-θ coupling-driven hierarchical workspace.

All of these assumptions take place within the context of the present traveling wave framework, with lower frequencies migrating objects across the cortex, which is particularly relevant to issues of maintenance and storage. Traditional 'standing' waves lead to periods when all neurons in a network are turned 'off', whereas traveling waves can ensure that sub-portions of a network remain consistently active (Bhattacharya et al., 2022), directly compatible with the mosaic-like architecture of posterior temporal and inferior frontal cortices in semantic integration processes (Murphy et al., 2022a).

## Causal Evidence

This section will explore current evidence that the neural architecture outlined here for ROSE may have some causal-explanatory power.

Beginning with the basic components of ROSE, we cannot directly claim that LFPs 'do' computation; their activity seems to index some underlying code that remains elusive in its formal character and neurochemical basis. Yet, aspects of brain dynamics have recently been associated with causal disruptions in cognitive and perceptual functions. A general guiding question that neurolinguists should consistently ask is the following: 'Is this signal important for the brain, or is it important for the neuroscientist?' There are many possible neural signals to record and make theoretical claims about, but a systemic means for adjudicating between correlation and causation remains a difficult obstacle. Nevertheless, some directions are opening up for establishing the causal role of oscillations.

As Snyder (2015) reviews, oscillations are increasingly being shown to play a causal, and not correlational, role in the perceptual segregation of sound patterns. Riecke et al. (2018) used speech-envelope-shaped transcranial current stimulation to conduct two experiments involving a cocktail party-like scenario and a listening situation devoid of any speech-amplitude envelope input. The results suggest effects on listeners' speech recognition performance, implying the existence of a causal role for speech-brain entrainment.

Moving closer to our concerns, Riddle et al. (2020) provide causal evidence from TMS that θ is involved in the control of working memory. That δ-β and θ-γ coupling are also causally involved in cognitive control (Riddle et al., 2021) suggests that phase-amplitude coupling can be used as a neurocomputational algorithm for feature extraction and association. Other research using hippocampal-targeted TMS-fMRI points to the causal role of θ-γ nested oscillations in memory encoding (Hermiller et al., 2020). θ has a causal role in spatial memory re-play (Zielinski et al., 2020). Fernández-Ruiz et al. (2019) looked at the rat brain and showed that learning and correct recall in spatial memory tasks are associated with extended sharp wave ripples. Artificially prolonging these ripples improved working memory performance, suggesting again a causal role for these rhythms in representational maintenance, at least over these specific neural populations. There also appears to be a causal role for α oscillations in perceptual binding (Zhang et al., 2019).

At the same time, in the realm of speech it has been suggested that 'forward entrainment' (neural entrainment that outlasts the entraining stimulus) "may instantiate a dynamic auditory afterimage that lasts a fraction of a second to minimize prediction error in signal processing" (Saberi and Hickok, 2022). Speech envelope tracking via θ may "leverage forward entrainment as a generative model of speech-segment timing to more efficiently process upcoming segments (where timing is inferred from expectations and priors in a Bayesian sense)" (Ibid.). This particular topic remains a contested space.

Regardless, as noted by Earl K. Miller, questions about the causal role of oscillations could also be asked of spikes. Both are signals that work together, and decoupling them seems difficult. Dismissing observed signals as 'epiphenomenal' without showing how, or why, does little to advance the neurobiology of cognition (Krakauer, 2022). Indeed, claims like 'X is epiphenomenal' are often synonymous with 'X does not fit my theory', as Miller has noted. It is also difficult to imagine what the alternative would be for the brain to implement complex computations and goal-directed behavior. Rhythmic fluctuations provide the foundations for these systems almost for free. Coordinating a large number of gates across a huge number of neurons seems impossible without this mechanism. That said, it is worth noting here that the frequency bands invoked in ROSE should not be seen as *types*, or fixed bounds for computation, but rather *tokens* of computation that are physiologically bound over brain maturation and development.

Following the chain of processes intrinsic to ROSE, I assume here that single-unit and assembly representations coding for discrete semantic and formal syntactic features would act as the spark igniting the neurocomputational architecture of the model. This would trigger LFP-scaled operations such as feature composition and sensorimotor transformations (grapheme-to-phoneme transitions, etc.), which in turn would be regulated by low frequency coordination of spike-LFP synchronization, and cross-regional synchronization via migrating waves of δ-θ coupling. An emerging consensus in this general direction is that spike-related activity is also present in low frequency activity, as well as in the γ band (Buzsáki et al., 2012; Łęski et al., 2013). Under ROSE, each scale of representation and computation can mechanistically be linked to the next. This departs from other current models that attempt, for example, to ground N400 ERP effects directly from single-unit responses, with no causal-explanatory mechanism linking them (Samimizad et al., 2022), or models that relate neurotransmitter profiles to the N400, without a mechanistic bridge tying together these scales of analysis (Hagoort, 2013).

## The Fallen Leaves Tell a Story: Prospects for ROSE

> "[W]e run the risk of being able to measure every cell (or subcellular component even) in the brain in a theoretical vacuum." (Pessoa, 2023, p. 358)

This article proposed a new neurocomputational architecture for natural language, termed ROSE (Representation, Operation, Structure, Encoding). Although the 'boxology' of classical models of language in the brain often placed emphasis on numerous aspects of the four components of ROSE, they did so in a way that did not narrow down the specific scale of system complexity, or rapid spatiotemporal dynamics necessary for a more neurobiologically plausible theory of syntax. Indeed, one of the common criticisms of neural oscillations is that they simply transfer localizationist claims from *regions* to *frequency bands*. Yet, what neural oscillations can provide is the potential to relate lower-level neural mechanisms, of the kind invoked by ROSE, to such frequency-specific activity. Although I have been concerned here with the "basic property" (Chomsky, 2021) of syntactic structure building (MERGE and its affiliate interface processes), future research should more concretely map complex syntactic processes, such as pronominal binding and *wh*-movement, to various stages of the interactions between the components of ROSE.

Importantly, while one researcher may be inclined to pursue accounts of relating generative grammar or minimalist parsing processes to ROSE, other researchers may argue for relations between the components of ROSE and other psycholinguistic models of language. It may turn out that minimalist grammar accounts, for example, have a more difficult time relating R with O, and O with S, than other accounts. For example, Stanojević et al. (2021) examine the derivations assigned by a near-context free formalism, Combinatory Categorial Grammar (CCG). CCG improves BOLD signal modeling in six language-relevant brain regions, and including a parsing step facilitating late-attachment of modifier phrases improves the fit in anterior temporal and inferior frontal cortices. Meanwhile, other researchers (Brennan et al., 2016; Li and Hale, 2019) discovered that node counts on X-bar structures derived by minimalist grammars predict unique variance in BOLD signal in the posterior temporal lobe.

Given that basic structure-building mechanisms are associated under ROSE as originating in posterior temporal cortex, and that extra-syntactic processing stages commonly implicate inferior frontal cortex, these joint results may suggest that different grammatical theories can capture varying aspects of sentence processing complexity (basic MERGE vs. late parsing stages and dependencies).

Some exciting prospects for testing ROSE come in the form of multi-channel recordings with broad cortical access using planar microelectrode arrays, implanted intracortically during awake brain surgery. These have recently been used to unveil traveling waves and also single-unit responses indexing basic mathematical operations in brain tumor patients in lateral parietal cortex (Eisenkolb et al., 2022). Both high γ power and the speed of traveling waves were greater when processing higher numerosities. Scaling from single-units to traveling waves would permit a comprehensive language mapping of the kind needed to test and refine ROSE.

Brain models informed by computational concerns will continue to be needed in the neurosciences, in particular as we approach the advent of widespread availability of single-unit human recordings (Rust and LeDoux, 2023). This (partially) 'outside-in' perspective has been critiqued recently. Buzsáki (2019) advocates for an inside-out perspective on building neural models of cognition, and considers the classical Marrian framework a purely outside-in perspective. Yet, Marr himself stressed that the three levels should be investigated in parallel, not necessarily prioritizing any given level (see also Marcus, 2001). But this is a balancing act: too much outside-in, and we will soon get claims about 'Neo-Davidsonian existential closure neurons'; too much inside-out, and we will receive calls to dismantle models of sentence parsing. Regardless of one's philosophical bent, single-unit and other types of intracranial recordings are plainly the most direct and reliable means to further test and refine ROSE (McCarty et al., 2022), in particular given recent independent assessments that spontaneous BOLD activity may be more closely aligned with offline plasticity and homeostatic processes than online fluctuations in cognitive content (Laumann and Snyder, 2021), and sceptical editorial commentary from major journals concerning the validity of fMRI: "[I]t is extremely difficult to conclude that functional connectivity as measured by functional MRI genuinely measures information exchange between brain regions" (Kullmann, 2020, p. 1045; but see also Esfahlani et al., 2020; Toi et al., 2022).

More broadly, ROSE is sympathetic to recent moves in philosophy of biology to view a range of biological constructs as processes rather than objects (Nicholson and Dupré, 2018). To invoke David Bohm (Bohm and Biederman, 1999, p. 12), "[t]here really is no 'thing' in the world" (see also Murphy, 2023b). ROSE is built entirely from neurocomputational mechanisms that are already known to subserve clusters of generic perceptual and cognitive operations. As Hasson et al. (2018) note, many language-specific interpretations of experimental neurophysiological and electrophysiological data might be better seen as implementations of highly generic processes, like monotonic integration of information, establishment of coherence, prediction, and representational binding. While Meyer (2018) reasonably speculates that cross-frequency coupling may be involved in "the neural binding of discrete phonological units at different granularity levels", under ROSE this generic process may be implicated in binding and associating other features ranging far beyond phonology.

Questions that remain concern the precise cortical infrastructure of ROSE, and how best to establish a mathematically rigorous algorithmic architecture. For example, superficial cortical layers oscillate at faster frequencies, and deeper layers at slower frequencies, potentially due to different interneurons (e.g., FS-PV vs. SOM) (Neske and Connors, 2016), while superficial layers are sources of feed-forward structural connections, and deeper layers are sources of feed-back connections (Mendoza-Halliday et al., 2022). At the same time, the relation between these remains unclear since increases in high $\gamma$ power are not always found alongside decreases in low frequency power; this depends on cortical region (Fellner et al., 2019).

## Conclusion

The development of the present ROSE architecture comes partly in response to the current absence of oscillatory phase coding in models of natural language syntax. This absence is growing increasingly stark given that this mechanism has long been known to be computationally useful in executing rapid neural binding. Many researchers who have kept to frameworks emerging purely from MEG-derived event-related fields, scalp EEG-derived event-related potentials, and fMRI-derived BOLD responses have begun to lose hope in the possibility of finding the neural basis of syntax, but have done so without moving beyond the limitations of their scale of analysis or recording, or linking hypotheses. For example, Pylkkänen (2019) writes that "[t]he neuroscience-of-language field has long assumed that our brains build syntactic structure during language processing. Today, it is reasonable to question this assumption".

Under ROSE, there is no sense to be made out of claims that syntax (a complex series of operations and representations) lies in one specific brain region. Even claims about syntax existing across multiple regions via white matter connectivity require substantial enrichment via the four components of ROSE. Each level of linguistic representation and structure-building requires its own suitable scale of neural organization, not just cartographic mapping. In this respect, one goal for the future would be to refine our understanding of the principled links between each of the four components of ROSE (i.e., how R mechanistically maps to O, and O to R), since this forms the crux around which the distinct scales of system complexity can be most directly tested and falsified. For instance, the line between 'oscillation or not' can sometimes be blurred, with a leaky integrate-and-fire model being both an oscillator and a model of evoked responses, and there is also heterogeneity in the possible mechanisms for oscillations (Doelling and Assaneo, 2021), and hence how they relate to single-unit information sources. At the same time, oscillations do in fact display some unique properties, such as their eigenfrequency, their Arnold Tongue, or their independence of rhythmic stimulus (their 'echo') (Fröhlich, 2015; van Bree et al., 2022). Many possible candidates for linking the four ROSE components exist once we refine our biophysical models of oscillations.

In terms of their roles in shaping information processing in cognition, the currently identified frequency-specific mechanisms seem to align with an emerging consensus that slow frequencies control input sampling, α/β gate information flow, and high frequency activity is controlled by slower rhythms (ElShafei et al., 2022).

Research into this topic may progressively validate the present assignment of linguistic constructs to particular scales of neural organization. Alternatively, it may ultimately transpire that the foundational structures of natural language syntax are not to be found at any of these scales, and will instead remain at a remote distance to modern science, lost to us on "the wide fathomless sea of living existence" (Montgomery 1886, p. 574).

# Appendix: Glossary of Terms

**Agreement:** When the form of a word/morpheme covaries with that of another word or phrase. Compare 'John runs to the park' with 'We run to the park', where the form of the verb is conditioned by whether the noun is singular or plural.

**Binding theory:** A set of principles accounting for the distribution of anaphoric elements (e.g., pronouns). A pronoun, or 'bindee', typically has an antecedent, or 'binder', as in 'John said he was happy', where the pronoun can successfully refer to the noun, unlike in '*He said John was happy'.

**Brodmann area (BA):** A region of the cortex defined by its cytoarchitectonics, or cell structure.

**C-command:** An expression of the relationship between nodes on a hierarchically organized syntactic tree. If a node has any 'sibling' nodes (nodes which are dominated by the same node) then it c-commands them, and if not then it c-commands every node that its dominating 'parent' node c-commands.

**Cell assembly:** A network of functionally connected neurons that is activated by a particular mental process and whose excitatory connections have been strengthened in time.

**Content word:** Words which name objects and their qualities. These are typically nouns but can also be verbs, adjectives and adverbs.

**Cross-frequency coupling:** When interactions between discrete frequency bands give rise to more complex regulatory structures. For instance, phase-amplitude coupling denotes the statistical dependence between the phase of a low-frequency band and the amplitude of a high-frequency band.

**Embedding:** The ability for a linguistic unit to host within it another linguistic unit.

**Erdős-Rényi network**: A random graph where each possible edge has the same probability $p$ of existing, and the degree of a node $i(k_i)$ is defined as the number of connections it has to other nodes. The degree distribution $P(k)$ of the network is a binomial distribution, decaying exponentially for large degrees $k$, permitting only small degree fluctuations. Facilitates coupling between slow frequency components.

**Function word:** Words denoting grammatical relationships between content words, such as prepositions, pronouns and conjunctions.

**Gain modulation:** How neurons nonlinearly combine information from multiple sources.

**Generative grammar:** The branch of linguistics which assumes that natural language is a mental computational system of rules generating an unbounded array of hierarchically structured expressions, with varying degrees of acceptability.

**Labeling**: The categorization of a MERGE-generated object at the point of conceptual interpretation, providing an asymmetric syntactic identity based on which element is most structurally prominent and easiest to search for, e.g., Z in WS = {Z, {X, Y}}.

**MERGE:** The computational operation which selects two objects from the lexicon, $\alpha$ and $\beta$, and forms an unordered set mapping them to an active workspace, WS = {$\alpha,\beta$}.

**Oscillation:** The unfolding of repeated events in terms of frequency. In the context of the brain, neural oscillations (or brain rhythms) are repetitive patterns of activity in caused by excitatory and inhibitory cycles in cell assemblies.

**Phase synchronization:** When multiple cyclic signals oscillate such that their phase angles stand in a systematic relation.

**Phi-feature ($\varphi$):** Linguistic features of Person, Number and Gender.

**Phonology:** The system of sound, or a set of sound-related features and rules stipulating how these features interact in a given language.

**Recursion:** The hallmark of natural language syntax; when a linguistic rule can be applied to the result of the application of the same rule, creating, for instance, 'nested' structures like 'John, who likes Sarah, will come to the party' from 'John will come to the party'.

**Syntax:** Informally termed the 'grammar', this is the set of principles governing the structure of morphologically complex word-like elements, phrases and sentences, and their combinatorial processes.

**Word movement:** A core concept in generative grammar whereby syntactic objects are displaced from the position where certain of their features are interpreted.

# Acknowledgements

My sincere thanks to all members of the Tandon Lab at the Department of Neurosurgery, UTHealth, who taught me to appreciate that what linguists wish to find in the brain is scarcely a realistic picture of what we can actually observe, including Nitin Tandon, Oscar Woolnough, Meredith McCarty, Kathryn Snyder, Tessy Thomas, Aditya Singh, Patrick Rollo and Kiefer Forseth. My thanks go to Patrick Trettenbrein for comments of an earlier draft. I would also like to thank Antonio Benítez-Burraco, Karl Friston, Koji Hoshi, Evelina Leivada, Gary Marcus and Jae-Young Shim for their more recent collaborative efforts with me, which have helped shape my understanding of the limits of the neurolinguistic enterprise and have each directly influenced the framing of at least one of the four components of the model presented here. This work was supported by the National Institute of Neurological Disorders and Stroke (Grant NS098981).